\documentclass[journal]{IEEEtran}
\usepackage{graphicx}
\usepackage{amssymb}
\usepackage{framed}
\usepackage{subcaption}
\usepackage{caption}
\usepackage{hyperref}
\usepackage[table,xcdraw]{xcolor}
\usepackage{dblfloatfix}    % To enable figures at the bottom of page

\ifCLASSINFOpdf
\else
\fi
\begin{document}
%
% paper title
% Titles are generally capitalized except for words such as a, an, and, as,
% at, but, by, for, in, nor, of, on, or, the, to and up, which are usually
% not capitalized unless they are the first or last word of the title.
% Linebreaks \\ can be used within to get better formatting as desired.
% Do not put math or special symbols in the title.
\title{Operational Learning-based Boundary Estimation in Electromagnetic Medical Imaging}
%
%
% author names and IEEE memberships
% note positions of commas and nonbreaking spaces ( ~ ) LaTeX will not break
% a structure at a ~ so this keeps an author's name from being broken across
% two lines.
% use \thanks{} to gain access to the first footnote area
% a separate \thanks must be used for each paragraph as LaTeX2e's \thanks
% was not built to handle multiple paragraphs
%

\author{A. Al-Saffar,
        A. Zamani,~\IEEEmembership{Member},
        A. Stancombe,~\IEEEmembership{Student Member},
        and~A. Abbosh,~\IEEEmembership{Senior~Member,~IEEE}% <-this % stops a space
\thanks{All authors are with the School of Information Technology and Electrical Engineering (ITEE), The University of Queensland, Brisbane, QLD, 4067 St.~ Lucia.}% <-this % stops
\thanks{Manuscript received November 30, 2020}}

\maketitle

% As a general rule, do not put math, special symbols or citations
% in the abstract or keywords.
\begin{abstract}
Incorporating boundaries of the imaging object as a priori information to imaging algorithms can significantly improve the performance of electromagnetic medical imaging systems. To avoid overly complicating the system by using different sensors and the adverse effect of the subject's movement, a learning-based method is proposed to estimate the boundary {\color{blue}(external contour)} of the imaged object using the same electromagnetic imaging data. While imaging techniques may discard the reflection coefficients for being dominant and uninformative for imaging, these parameters are made use of for boundary detection. 
% Numerous design alternatives were evaluated and juxtaposed with the overarching goal of reaching a reliable operational model in clinical settings. 
The learned model is verified through independent clinical human trials by using a head imaging system with a 16-element antenna array that works across the band 0.7-1.6 GHz. The evaluation demonstrated that the model achieves average dissimilarity of 0.012 in Hu-moment while detecting head boundary. The model enables fast scan and image creation while eliminating the need for additional devices for accurate boundary estimation. 
\end{abstract}

%TODO add a case where there is a single erroneous estimation to be fixed.
% Note that keywords are not normally used for peerreview papers.

% For peer review papers, you can put extra information on the cover
% page as needed:
% \ifCLASSOPTIONpeerreview
% \begin{center} \bfseries EDICS Category: 3-BBND \end{center}
% \fi
%
% For peerreview papers, this IEEEtran command inserts a page break and
% creates the second title. It will be ignored for other modes.
\IEEEpeerreviewmaketitle

\section{Introduction}
% The very first letter is a 2 line initial drop letter followed
% by the rest of the first word in caps.
% 
% form to use if the first word consists of a single letter:
% \IEEEPARstart{A}{demo} file is ....
% 
% form to use if you need the single drop letter followed by
% normal text (unknown if ever used by the IEEE):
% \IEEEPARstart{A}{}demo file is ....
% 
% Some journals put the first two words in caps:
% \IEEEPARstart{T}{his demo} file is ....
% 
% Here we have the typical use of a "T" for an initial drop letter
% and "HIS" in caps to complete the first word.
\IEEEPARstart{E}{lectromagnetic} (EM) medical imaging in the microwave band is a relatively new modality, finding applications in the detection and characterisation of breast cancers \cite{bond2003microwave,Klemm2009,Bassi2013}, ischemic and haemorrhagic strokes \cite{Semenov2008,Scapaticci2012,Ireland2013}, and torso imaging applications \cite{Celik2014,Rezaeieh2014}. The physical phenomenon that enables this modality is that the dielectric permittivity and conductivity of human tissues vary considerably over the microwave frequency band, particularly between healthy and unhealthy tissues. After transmitting and receiving signals through the body using antennas in known positions with known properties, an inverse scattering problem might be utilized. The solution of this problem can yield an image of the dielectric properties of the body under test, and hence the position of different tissue regions, allowing for disease localisation.

The imaging techniques currently reported in the literature typically require a priori information about the propagation or scattering model of the imaging domain. The dire need for prior information stems from the fact that solving inverse scattering problem is an expensive computational process that requires the estimation of a large number of unknowns (dielectric properties of tissues). 
{\color{blue}
In \cite{entry_point} the authors suggest estimating `entry points' of signal into the object to aid a decluttering mechanism.
}
Introducing the boundary of the imaged object to the imaging algorithm can enhance its detection accuracy by providing a more accurate propagation model. Moreover, it can reduce the computational complexity of the problem by reducing the dimension of the search space. 
{\color{blue}
Last but not least, the vast majority of imaging techniques in biomedical settings use some `decluttering' mechanism to focus on a target that is typically the pathology \cite{cancer}. This decluttering process physically corresponds to removal of some aspect of the scanned object, e.g. effect of skull reflections in head imaging. As such, the output image only shows the anomaly and lacks the boundaries of the imaged object. This is not palatable to clinicians as it lacks context \cite{stroke1, stroke2}. Thus, object boundary finds its place as one aspires to construct a holistic image as a final outcome.
}
The collection of boundary information in a clinical environment where the subject's movement is not controllable is a difficult process. Using laser sensors \cite{williams2011laser} is the most accurate strategy for the detection of the object's surface in current literature. However, the laser scan should be performed at the same time as the EM scan to prevent generating an incorrect propagation model due to the inevitable subject movements during the scans. Furthermore, embedding such sensors to the EM imaging apparatus complicates the system. To overcome such drawbacks, the same EM data captured for imaging must be utilized to detect the boundary. Using EM data offers a low-cost, convenient, and reasonable alternative to complicated sensors embedded in any EM imaging system.

Utilising the EM data to find the boundary of the imaged object has been researched in the literature, particularly in the context of breast imaging \cite{Williams2006,Williams2008,Winters2008,Li2005b}. In \cite{Williams2006}, deconvolution of the time-domain reflected signals was used to approximate the impulse response of the scattering interface.
% and provided simulation-only results to verify its efficacy.
The skin thickness was also estimated with this technique.
In \cite{Williams2008} the same technique further utilized to generate full volumes of a breast under image in presence of multiple antennas surrounding the object. \cite{Winters2008} also generated breast volumes using the time-domain reflected signals. 2D matched filters are made use of to approximate the time delay of the signal to the skin  \cite{Li2005b}. More recently, \cite{Kurrant2017} detailed an approach for approximating the breast surface using sparse data collected from either microwave or laser sensors.

% \section{Motivation Learning Models}
% \label{motivation}
The aforementioned approaches implicitly assume that dynamics of the problem are linear and resort to system identification techniques and simulations to tackle the problem.
% To inspect the potential of such approach, a small bank of 40 simulations was built. In these simulations, the object of interest was a simplified circular phantom (a 5-tissues head model) with properties equivalent to the average of human head, and is placed at predefined distance from the antennas. In each simulation of this bank, the phantom is set with a distance of 1 mm increment over the previous one, starting from 1 mm ending at 40 mm, which is the range of interest. The final result is a tensor of S-parameters of shape $40 \times 1011 \times 16 \times 16$. From this point, various transformations and decompositions were applied to the individual tensors. 
While the theoretical literature lacks results that explicitly include a boundary-related variable in its formulation, it is not a hard task to experimentally ascertain that problem dynamics are non-linear. E.g. a small bank of simulations featuring incremental distance between scaterrer and imaging array can provide insight into the underpinning problem. From the dataset accumulated, and any transformation from Dynamic Mode Decomposition \cite{schmid2010dynamic}
% Non-Linear Fourier Transform \cite{FNFT2018} 
or Scattering Wavelet Transform \cite{bruna2013invariant}, can be applied while the behaviour of resulting coefficients over different simulations featuring incremental distances can be observed. Invariably, the coefficients exhibit complicated behaviours and changing trends.

% We opt to exclude any results from these simulations for the followign reasons:
% 1- This is an intro section.
% 2- The resulting coefficients are large in size and only a GIF can give them due presentation.
% 3- it is not a controversial claim made here that necessitates a proof. Everyone agrees with the fact states. Lastly, this has little if any impact on comprehension of the work presented.
% \cite{Kossaifi2019TensorLyTL, sidiropoulos2017tensor}, Dynamic Mode Decomposition (DMD) \cite{demo18pydmd, schmid2010dynamic} , Non-Linear Fourier Transform (NLFT) \cite{wahls2015fast, FNFT2018} and Scattering Wavelet Transform \cite{Andreux2020KymatioST, bruna2013invariant}
% It became apparent that no simple feature can be exploited to solve the problem in a direct way. 
% \footnote{Animations of simulated bank of data undergoing the transformations are provided supplementary materials.}.
% In an equally straight-forward manner, one can extract features from the raw signals (or its transformations) such that this feature correlates with the distance. 

\begin{figure}[!t]
    \centering
    \includegraphics[width=0.48\textwidth]{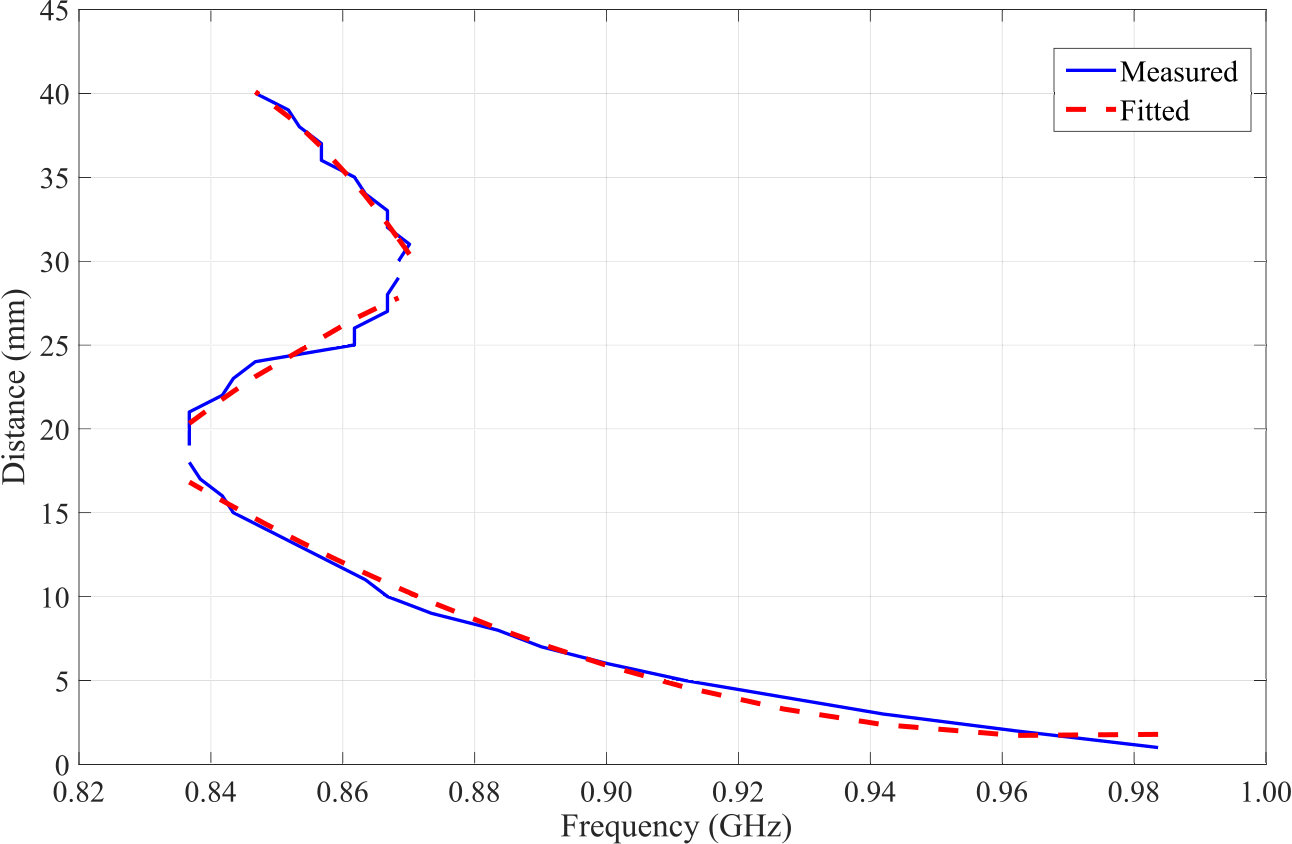}
    \caption{\small Overlapping trends of resonance shift. The first consistent partial curve goes from 0 -- 18 mm, while the second is acting in the interval 18 -- 30 mm and the last one is active in the regime 30 -- 40 mm.}
    \label{fig:resonance_trend}
\end{figure}

An equally straightforward and partially successful approach from the literature is Resonance Shift \cite{zamani2017boundary} which is based on manual feature extraction from the scattered signal in the frequency domain. In essence, the technique utilizes the observation that antenna load affects the resonance frequency. The fundamental limitation however, stems from changing trends of this feature as the distance changes, crippling its usefulness to limited range of distances where the trend is consistent, beyond this range, the bijection is lost and the resonance is resolved to up to 3 possible distances. The trends and the ranges are shown in Figure~\ref{fig:resonance_trend}.

% The trend exhibited by resonance shift does not seem to be an arbitrary artifact, many other features examined showed similar behaviour, illustrating how deeply this trend is rooted in the underlying phenomenon. 

A potential fix of the drawbacks of this technique by complementing the frequency resonance feature with other features that could help to resolve the ambiguities. E.g. a feature that shows similar curve but has different turning points. Towards this goal, various spectral, temporal and statistical signal features were examined for this purpose \cite{barandas2020tsfel}. While many such features can be found, none proved to be robust enough to forge a reliable algorithm. Resonance frequency has immense immunity against noise, making it a particularly useful feature. In comparison, features that are based on magnitude of signal are susceptible to noise and as such, will not be suitable in real-life settings despite potentially performing well in simulation settings. 
% It is worth mentioning that many signal features exhibit a behaviour similar to resonance shift depicted in \ref{fig:resonance_trend}, e.g. total variation, spectral kurtosis, illustrating how deeply rooted is this trend in the underlying phenomenon. 

Given the observed behaviour of manually extracted features, it comes to light that Machine Learning (ML) stands apart as a promising candidate technique for tackling the task. This work proposes a learned model for inferring the boundaries from the reflection coefficients of the signals captured. To train the proposed model, a feasible laboratory mechanism is setup to provide surrogate dataset essential for training. The trained model then must have the ability to generalize from a \textit{single} experimental phantom to arbitrary real human heads, which are coveted subjects of interest in the context of this work.

The remainder delves into the intricacies of the proposed learning-based design. Section \ref{sec:problem_setup} describes problem setup and the data compilation mechanism. Then, model design is presented in section \ref{sec:design}. Section \ref{sec:results} describes model training, ground-truth procurement procedure and showcases the clinical results. Finally, Section \ref{sec:conclusions} discusses those results and draws conclusions.

% \section{Proposed method}
% To estimate the boundary of the imaged object by reflection coefficients of the EM signals, a multilayer-perceptron model is designed and trained with a rich realistic dataset.
% The proposed model predicts the normal distance between the antennas of the imaging array and the outer layer of the imaged object.
% The discrete normals are then made into points, which in turn are interpolated to generate the final boundary.
% The model is fit with training dataset that is built in a laboratory setup comprising a 16-antenna array system and a head phantom that is controlled by a CNC machine. 
% Since the inference is made per antenna, the model can generalize to arbitrary shapes, though the training dataset includes only one phantom. As such, the model remains oblivious to the overall shape whether at training time or inference time when the subject of interest is a human. 
% To ascertain that the model learns the underlying physical phenomenon necessary to generalize from phantom to human subject, the model is kept small with only 379 parameters. 
% The reflection coefficients measured over 451 frequencies are reduced to 10 using Principal Component Analysis (PCA). This reduces the size of input  and facilitates smaller models that are less prone to overfitting plague. The overall scheme is depicted in Figure~\ref{fig:scheme}.

% perm_cond
\begin{figure}[!t]
    \centering
    \includegraphics[width=0.45\textwidth]{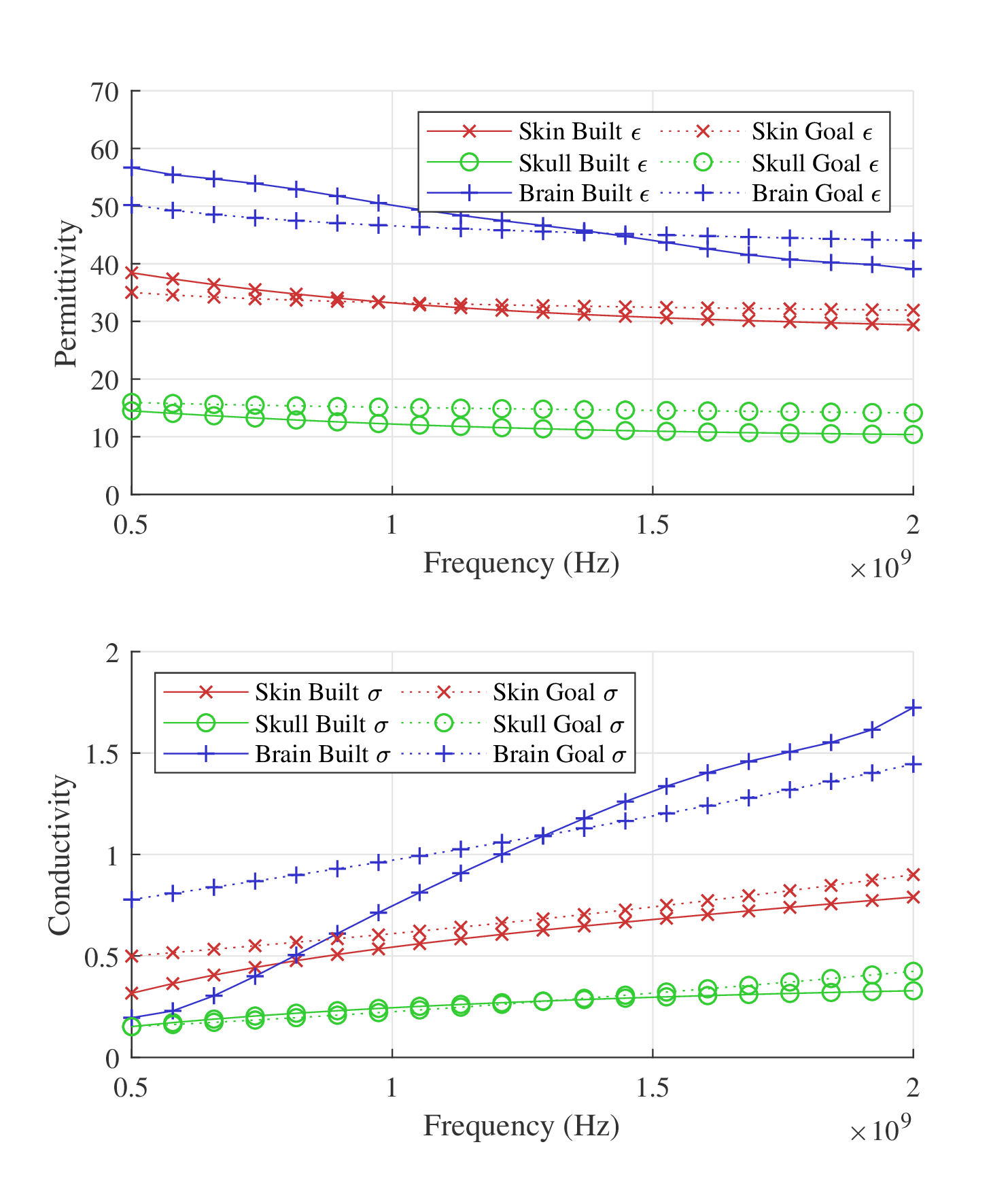}
    \caption{\small Permittivity and conductivity of the phantom layers, where solid lines are the manufactured properties and dotted lines are the desired properties}
    \label{fig:perm_cond}
\end{figure}

\section{Problem Setup}
\label{sec:problem_setup}
% \label{sec:dataset}

It is almost invariably the case that in many real-world problems, when a learning-based approach is adopted, data becomes an intrinsic extra dimension of complexity added to problem description. In particular, compilation of a sufficiently large dataset for training is prohibitively expensive. 
% As such, dataset collection process is now part of problem description and defines to a great extent. 
To overcome this, the human head is mimicked with a three tissue realistic phantom.
% pictured in Figure~\ref{fig:experimental_setup}.
The three dielectric layers are: a 7 mm thick solid outer skin / fat / muscle average layer constructed from a 34.6\% epoxy, 15.4\% graphite, 46.2\% alumina, and 3.8\% brass powder mix ratio by weight; a 6 mm thick solid inner skull average layer constructed from a 55.9\% epoxy, 0.3\% carbon black, and 43.8\% alumina powder mix ratio by weight; and a hollow inner region filled with brain tissue emulating fluid, made up of a homogenised 60\% glycerol and 40\% water mix ratio by weight. The dielectric properties of these manufactured layers and also of the tissue averages that informed their construction (derived from \cite{Gabriel1996}) can be seen in Figure~\ref{fig:perm_cond}. It is apparent that the solid layers are better matched to the goal dielectrics than the liquid layer, but the liquid layer still performs acceptably while allowing for easy insertion and movement of a disease emulating target within the phantom. The mixture is non-toxic and inexpensive, making it convenient for use in a lab environment.

% antenna array
\begin{figure}[!t]
    \centering
    \begin{subfigure}[]{0.22\textwidth}
        \includegraphics[width=1.\textwidth]{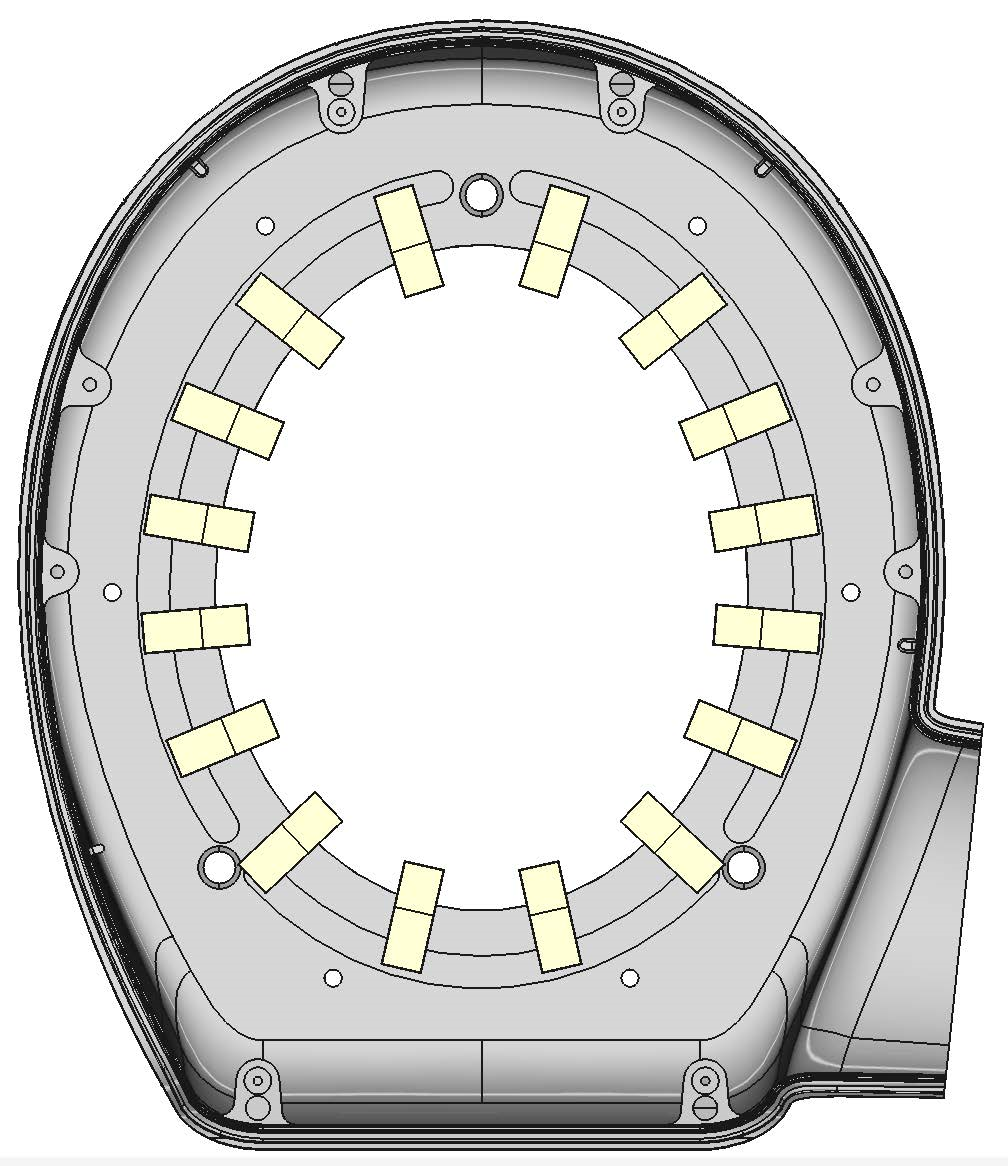}
        \caption{}
        \label{fig:antenna}
    \end{subfigure}
    \begin{subfigure}[]{0.22\textwidth}
        \includegraphics[width=1.\textwidth]{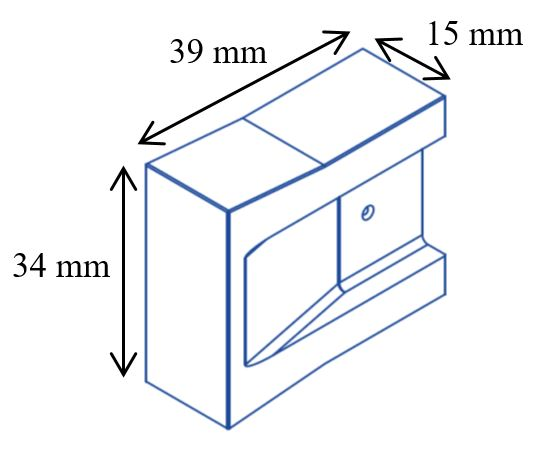}
        \caption{}
        \label{fig:array}
    \end{subfigure}  
    \caption{\small (a) Antenna array made of 16 tapered filled waveguides shown in (b)}
    \label{fig:antenna_array}
\end{figure}

\begin{figure}[!t]
    \centering
    \includegraphics[width=0.45\textwidth]{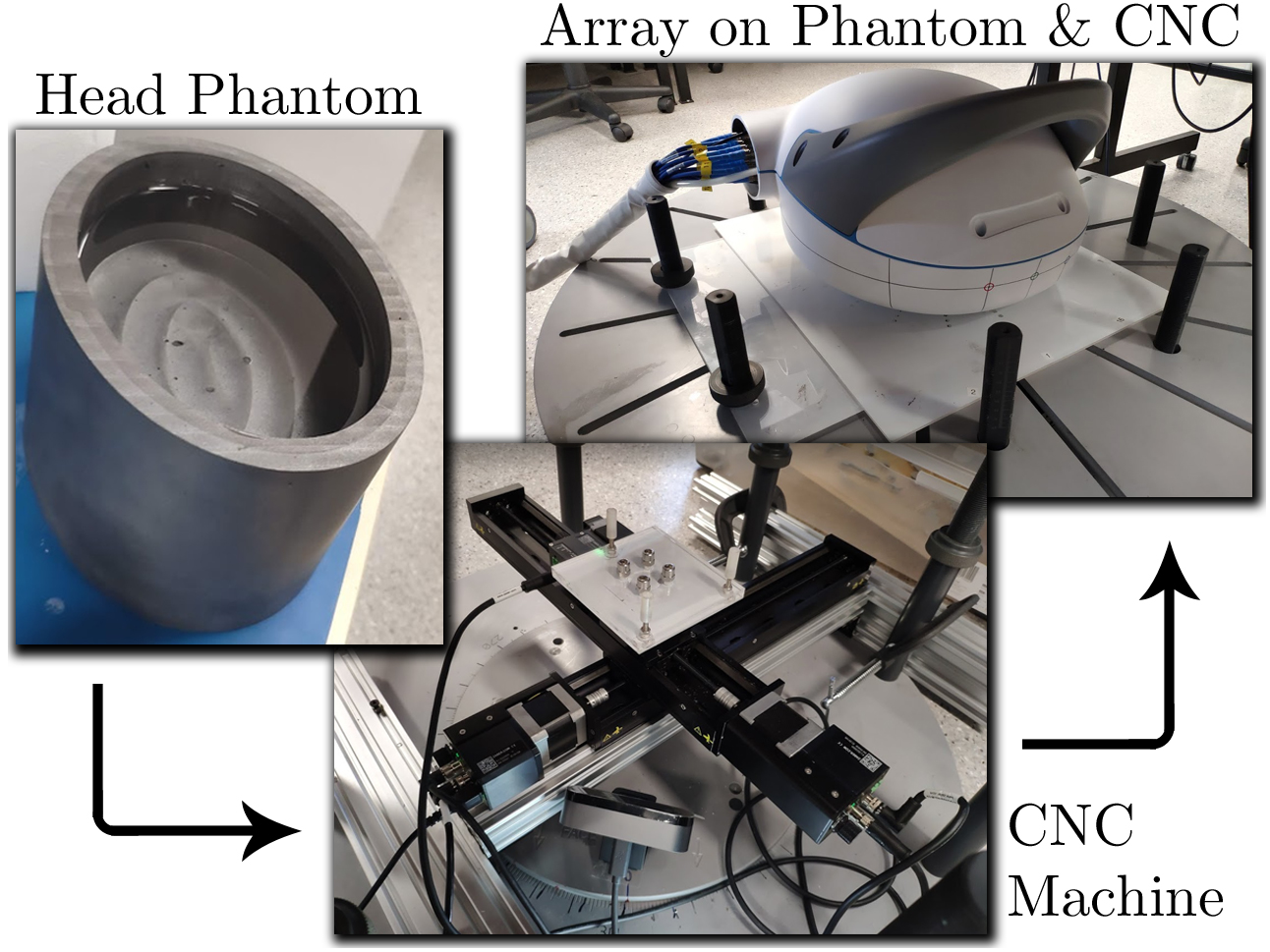}
    \caption{\small Experimental setup -- the head phantom was placed upon the CNC machine, and the antenna array was then placed upon the head phantom}
    \label{fig:experimental_setup}
\end{figure}

Experiments were collected using a 16-element antenna array shown in Figure~\ref{fig:antenna_array}. The utilized antenna is a tapered ceramic filled waveguide with dimensions of $15 \times 34 \times 39 {mm}^3$ and ceramic dielectric properties of permittivity $\epsilon_r = 45.5$ and $\sigma = 0.01$. This enables the EM wave matching to the head as well as antenna miniaturization. 
The head phantom in 444 random positions within the imaging domain. These positions were achieved through the use of a three-axis Computer-Numerically-Controlled (CNC) machine. The first two axes enabled lateral movement within the domain and the third axis provided rotation of the phantom. 
This experimental configuration is depicted in Figure~\ref{fig:experimental_setup}.

For each of the phantom positions within the domain, a full 16 port S-parameter matrix was captured. This was achieved through the use of a Keysight M9800A multiport Vector Network Analyser (VNA). The data was captured over the frequency band of interest, 0.7 -- 1.6~GHz, in 451 steps (2~MHz spacing). Calibration was applied to the data to minimise the effects of the cables and to move the measurement plane up to the input of the antennas.

%For the purpose of brevity, in the sequel, the data collected with the setup described in this section will be referred to as \textit{SSB data} (this is the initialism of Skin Skull Bucket), , sometimes referred to as real data in which humans are the subject of interest. SSB 

 An obvious discrepancy between training laboratory data and the real clinical data is that the former is compiled with phantoms while the latter come from real human heads as scattering objects. However, beyond this, the other major identified differences between realistic data and phantom data are:
\begin{enumerate}
    \item The setup in the laboratory is horizontal as depicted in Figure~\ref{fig:experimental_setup}. In the clinical environment however, the setup is made vertical as will be shown later. The vertical setup will cause the head to push with its weight downwards while in the horizontal setup this pressure is equal around the membrane. %Consequently, there will be much smaller distances from antennas in vertical setup.
    \item Hair is not taken into account in the phantom.
\end{enumerate}

These discrepancies disallow even performance of neural net across different datasets. Moreover, the training data recorded a smallest distance from any antenna of around 3.8 mm. The largest distance was around 17.5 mm. In reality however, distances are expected to fluctuate as wildly as small as 1 mm up to 20 mm. A 1 mm distance is likely to be encountered due to the aforementioned weight issue in vertical setup. In the opposite direction, there can be a scenario of small head (smaller than the single fixed size phantom) leading to large distances. Signals corresponding to such distances are \textit{unseen} in the training set. This constitutes one major challenge in the proposed learning-based model. Ideally, one aspires to see the neural net with the ability to generalize to ranges that are beyond the training dataset collected in laboratory settings. This simply boils down to genuinely learning the underpinning physical phenomenon as opposed to memorization which is the infamous plague of deep learning.
% Special techniques are further required to bridge this discrepancy.

The data described here is to be used for training and will be referred to in the sequel as \textbf{phantom data}, \textit{not to be confused} with clinical data which will be used for evaluation and is referred to as \textbf{real data}.

\section{Design}
\label{sec:design}

To present a well-rounded description of the proposed model, the section is organized into three modules, each of which addresses one independent component of design, delineating with reasoning and conclusions every choice made. Since the nature of data available for training a model determines to a great extent the architecture of that model, thus, part \ref{sec:ip_sig} starts by discussing how to handle the input signal from the system. Part \ref{architect} presents the proposed architecture. Lastly, \ref{resp_var} shows how to pick appropriate response variable to be predicted.

% experimental_setup

\subsection{Input signal}
\label{sec:ip_sig}
The scattering parameters measured are of shape $451 \times 16 \times 16$ where 451 is the number of frequencies. Feeding this size of data to a model is not optimal as it necessarily implies an over-bloated neural net and indeed proved inferior to selecting the most relevant parameters. In particular, the elements on the main diagonal that correspond to reflection coefficients provide information about the antenna itself (rather than its relation with the others) and lend themselves naturally to boundary estimation. Thus, the effectively used data is of shape $451 \times 16$. 
% {\color{blue}
Admittedly, the immediately adjacent S-parameters $S_{i, i\pm1}$ capture relevant information in this regard, however, 
{\color{blue}
this triples the input size which entails a larger model to consume the data. Larger models certainly have increased learning capacity, however, the aim is to learn a generic principle behind the phenomenon \textit{without} the peculiarities that are specific to phantom data (which is not applicable to the domain of interest; the real human data). To achieve this, irreducibly compact models are sought after as will be explained further subsequently.
}
% . A compact model performs decently on phantom data but is superior
% }
% Unless a separate auxiliary model works by consuming the adjacent transmission parameters, then the choice of including adjacent parameters only serves to bloat the model, and substantially increases over-fitting. 

% TODO Alternatively, one can represent shape as a sequence of data. Thus one can obtains arbitrary desired resolution of prediction, and, can enforce uniformity constraints easily by using total variation, and lastly one can even make use of symmetry which manifests itself as sinusoidal curve. One can use Fourier series and apply it to each half of the wave and use reconstruction loss as regularize of symmetry.

% % scalogram
% \begin{figure}[t!]
%     \centering
%     \includegraphics[width=0.45\textwidth]{images/scalo.png}
%     \caption{\small Scalogram conversion of one S-parameter signal. It is noticeable here that scalogram image is dull and has very little information encoded in it. Looking back at the signal itself, it is monotonic to a degree, and is not even nearly as rich as a speech signal where this technique is common. Using this approach is analogous to cracking a nut with a sledgehammer; data size is unnecessarily multiplied and little if any more information is revealed.}
%     \label{fig:scalogram}
% \end{figure}

\subsection{Architecture}
\label{architect}
Having made the choice of selecting reflection signals only, the next step is investigating suitable architectures for this input. Currently, two ubiquitous approaches exist in the literature to handle {\color{blue}
`time series'} data at hand. 

\textbf{\textit{The first}} major approach is using recurrent neural nets \cite{crnn_ecg}. More or less, this approach entails no preprocessing at all as this architecture is specifically designed for handling {\color{blue}
time series} data. 
% The use of this architecture is associated with inherent downsides that will be explained in the next part.
% RNNs come in various flavours, most of which were experimented thoroughly.

\begin{figure}[!t]
    \centering
    \includegraphics[width=0.4\textwidth]{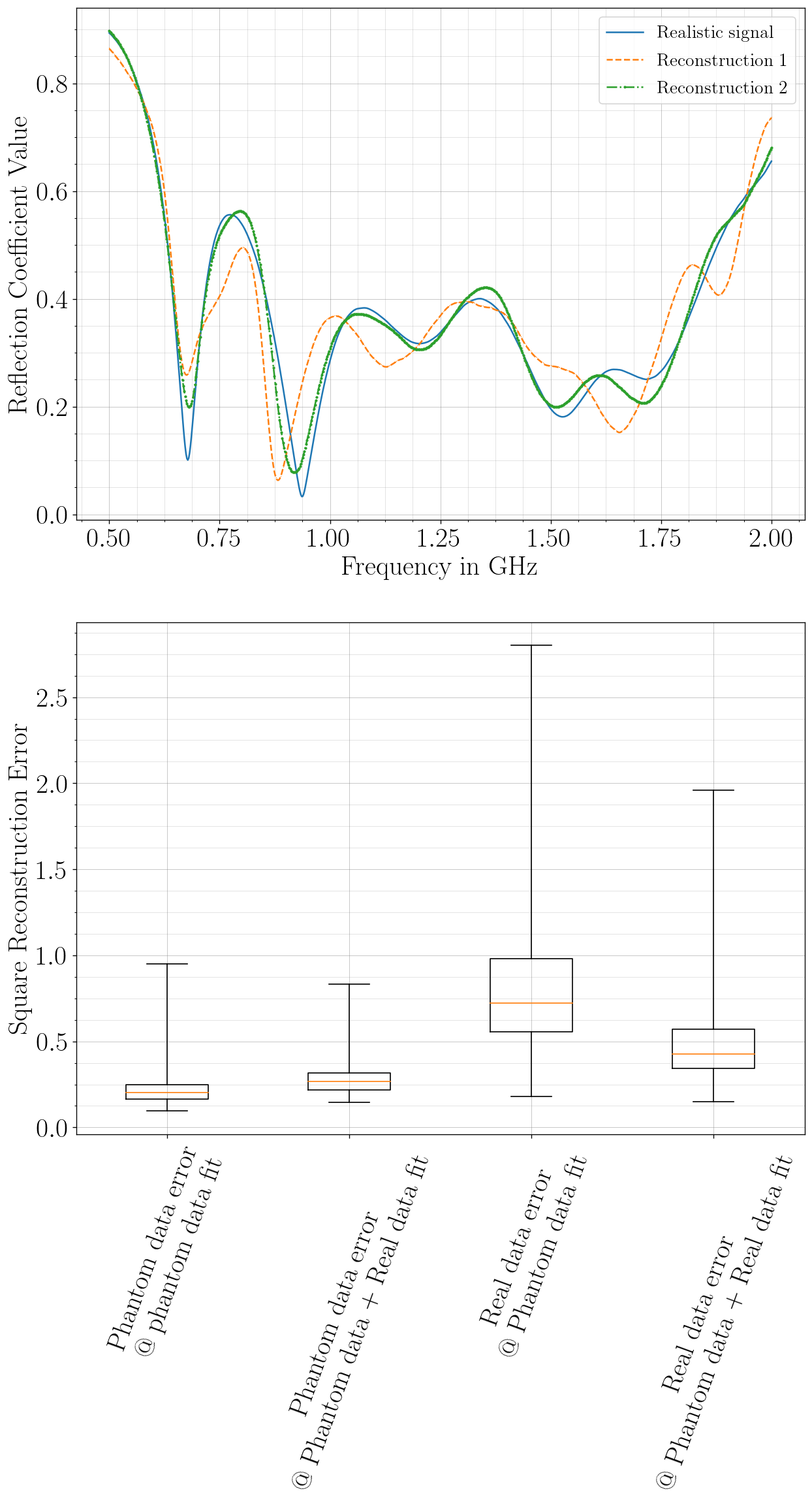}
    \caption{\small Top: A sample of clinical signal (out of training set) along with its reconstruction from its embedding for two scenarios; `Reconstruction 1` is due to PCA fit with training data, while `Reconstruction 2` is due to PCA fit with training data cross-fertilized with clinical data. Bottom: error bars for two fits on two data sources. While not perfect, the fit with padded real data achieves a decent square reconstruction error of 0.48 compared to 0.88 for real data.}
    \label{fig:recons_pca}
% \vspace{-6mm}
\end{figure}

\textbf{\textit{The second}} is to extract features from the {\color{blue}
input signal} using a host of transformation ranging from Spectrograms \cite{spectro_paper}, Scalograms \cite{scalo_paper} or Constant-Q transforms \cite{constq_paper}. All of those approaches perform a similar function in essence, that is convert signal to an image, then use a Convolutional Neural Network (CNN) which is famous for its efficacy. 
% An exemplar experiment with such approach is shown in Figure~\ref{fig:scalogram}.
% While the scalograms still exhibit healthy variations across different inputs
% \footnote{Animation is provided in the supplementary material.},
These approaches were proven effective for speech signals, biomedical signals that have rich information encoded in them. Such transformations help to manifest the information in image format that is conducive to CNN operation. That said, for smoother signals like the one in question, such transformation only leads to unnecessary data size multiplication while no more information is revealed with the new representation.
A more effective approach to process it is to summarize to few features and feed it to a small neural net. To summarize the signal, Principal Component Analysis (PCA) \cite{pca_paper} is used for reducing the dimensionality from effectively 451 down to 10. This size of input naturally results in a very thin neural net with good potential for learning the underlying physical phenomenon. This summarization approach however comes with its own drawback, being a data-driven technique, its performance is uneven across all signals. Namely, when PCA is fit on training data, its performance degrades when evaluated on unseen signals related to real humans. This degradation can be estimated by inspecting the reconstruction loss. If the reconstruction loss is high, then there is little hope of expecting the model to perform decently as there will be garbage in, garbage out scenario. One famous solution from the literate that lends itself naturally to PCA is Transfer Component Analysis \cite{tca} which is a domain adaptation technique aimed at finding principal components that are common between the source domain and the target domain. A more recent technique is Subspace Alignment \cite{fernando2014subspace} which aims at projecting the source and target domains to a common subspace. Both of those approaches achieved sub-optimal results as projection into common spaces leaves little of the information essential for identifying the label. A simpler trick involving the inclusion of clinical data with training set, proved to be more effective. Figure~\ref{fig:recons_pca} depicts the found results. 

{\color{blue}
The added clinical data is comprised of 36 measurements of volunteers. Thus, with 16 signals per measurement captured by the array, the auxiliary data adds up to 567 signals. Compared to the size of phantom data (7104), it constitutes 8\% of the total data. It is ought to be emphasized here that added clinical data is solely for the purpose of fitting PCA model, but not the neural network, as the clinical data has no labels.  
}

% pca loss

% response_options
\begin{figure*}[!t]
    \centering
    \begin{subfigure}[b]{0.3\textwidth}
    \centering
    \includegraphics[width=\textwidth]{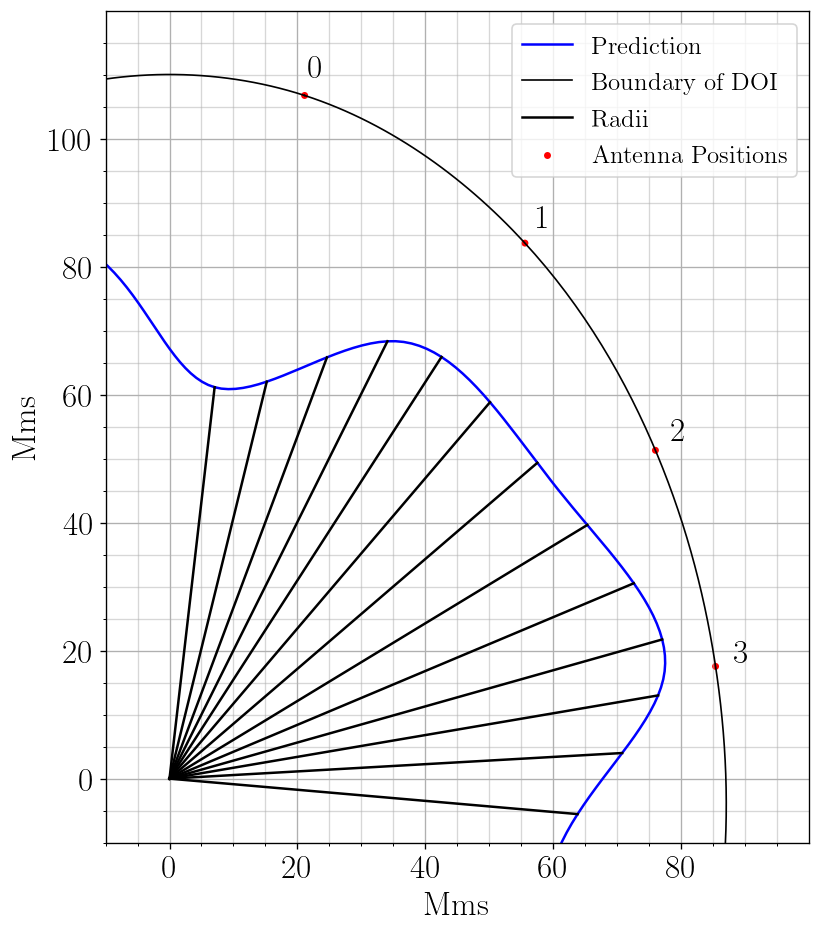}
    \caption{}
    \label{fig:radii}
    \end{subfigure}
    \hfill
    \begin{subfigure}[b]{0.3\textwidth}
    \centering
    \includegraphics[width=\textwidth]{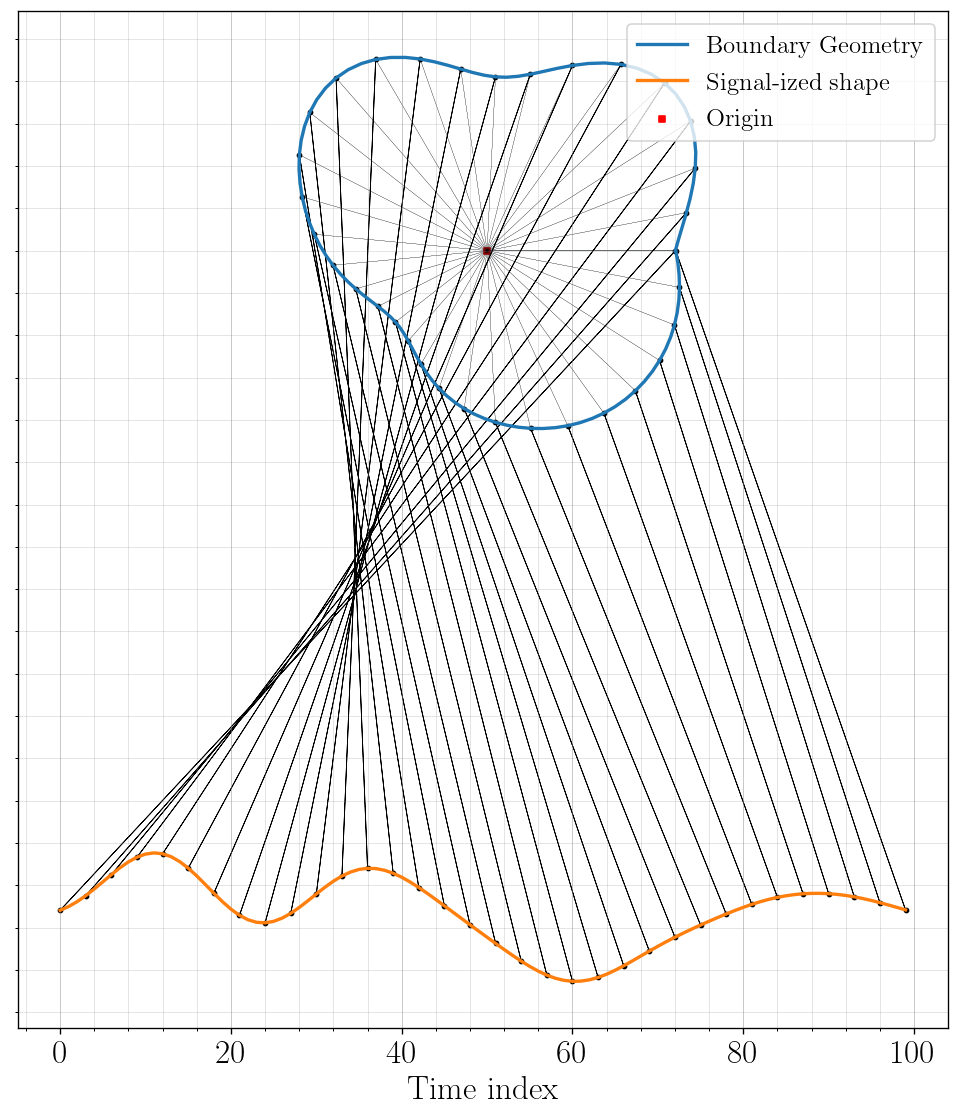}
    \caption{}
    \label{fig:signal}
    \end{subfigure}
    \hfill
    \begin{subfigure}[b]{0.3\textwidth}
    \centering
    \includegraphics[width=\textwidth]{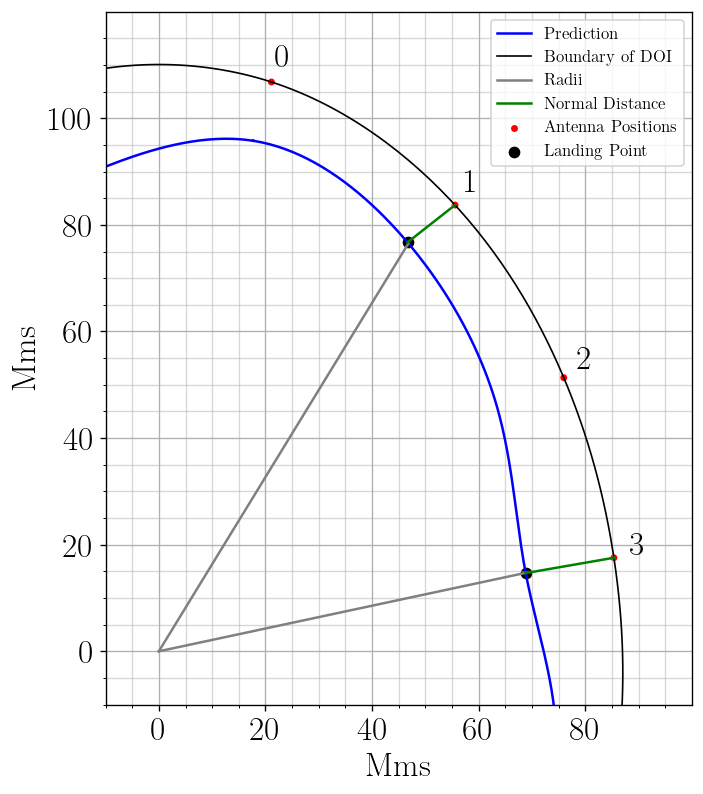}
    \caption{}
    \label{fig:normals}
    \end{subfigure}
    \caption{Response variable schemes. (a): Regressing the radii constituting the curve sampled at fixed angular rate. (b): Unwrapping boundary shape into a signal. (c): Depiction of the normals to DOI boundary (the direction of EM wave travel, shown in green). The corresponding radii for each normal is also depicted (black) to explain away the uneven performance in the independent multi-model paradigm; $model_{1}$ associated with antenna 1 exhibits poorer performance compared to $model_{3}$ which encounters a simpler geometric transformation from the normal to the radius.}
    \label{fig:response_options}
\end{figure*}

\subsection{Response variable choice}
\label{resp_var}
% This section exhausts formulation alternatives for boundary estimation.
% \subsubsection{Regression of the boundary}
To predict the boundary with a learning model, there exist a variety of formulations for the response variable to be estimated. Below, the pros and cons for these choices are examined.

\textit{Approach 1:} Direct regression of x-y coordinates of the boundary. %While the input being the entirety of S-parameters.
This design choice is heavy in the output. To cut the predictions in half, one can alternatively regress the radii of the points along the boundary separated by fixed angular steps as depicted in Figure~\ref{fig:radii}. This approach suffered from a major drawback; given that the neural net is required to predict the entire shape \textit{simultaneously}, it tends to memorize the shapes in training set. The training set is comprised of only one ellipsoid shape, shifted to various locations as explained in Section~\ref{sec:problem_setup}. Subsequently, the model fails to generalize to other shapes altogether.

% \subsubsection{Regression of the radii}
\textit{Approach 2:} To counter the drawback of the previous approach, an array of 16 neural nets can be utilized; each one is tasked with predicting the boundaries facing one antenna. As such, the boundary itself is split into 16 parts, and each model is required to predict one part, which is composed of several points. Ideally, this scheme is endowed with great flexibility and as such, can handle arbitrary shapes given that predictions are made independently for each antenna. In practice however, irregularities in the overall shape starts to occur. Concretely, this choice constitutes the diametric opposite the previous design and has extremely loose structure of predictions.  
% Additionally, neural nets performance was observed to be \textit{uneven}. Figure \ref{fig:normals} is illustrating the problem; some neural nets, e.g. the one corresponding to antennas 3 ($model_{3}$) performed better than $model_{1}$. While NN in theory have the capacity to learn arbitrary functions, $model_{1}$ is having a more difficult task regressing the radius at an angle. In comparison, for neural nets like $model_{3}$, the radius at that angle is in straight line with the aperture of the antenna, thus it is easier to regress that radius from the input.

\textit{Approach 3:} this is similar to the previous formulation but goes one step further by explicitly expressing the boundary as a signal and use models to predict signals. This conversion is depicted in Figure~\ref{fig:signal}. The approach is known in the signal processing community \cite{sigalize} and has also been carried over to data-driven models like Recursive Neural Nets \cite{tan2019time}, \cite{tan2020time}. That said, it is still suffers from the same downsides as approaches 1 and 2.

\textit{Approach 4:} This formulation fixes all the previous drawback by switching from regressing the radii, to regressing the normal lines from antennas to the boundaries as depicted in Figure~\ref{fig:normals}. Thus, to reconstruct the boundary, one can travel from antenna aperture point, by as much as the normal length, to arrive at a landing point. Next, the landing points are collectively interpolated to construct the final shape.

Additionally, with this approach, only one neural net is utilized to perform the task instead of 16 as in \textit{Approach 2}. Admittedly, allocating one model per antenna is a good choice as each antenna has slightly different physics. This arises from the fact that the antennas are not arranged in a circle, resulting in different mutual coupling for each one of them, different responses, etc. Thus, the choice made of having a single model instead of 16 seems to be inferior. However, this consideration is dwarfed by the benefits brought about by a single model acting on all antennas. Below is an enumeration of the advantages of the final proposed architecture:

% scheme
\begin{figure*}[!t]
    \centering
    \frame{\includegraphics[width=\textwidth]{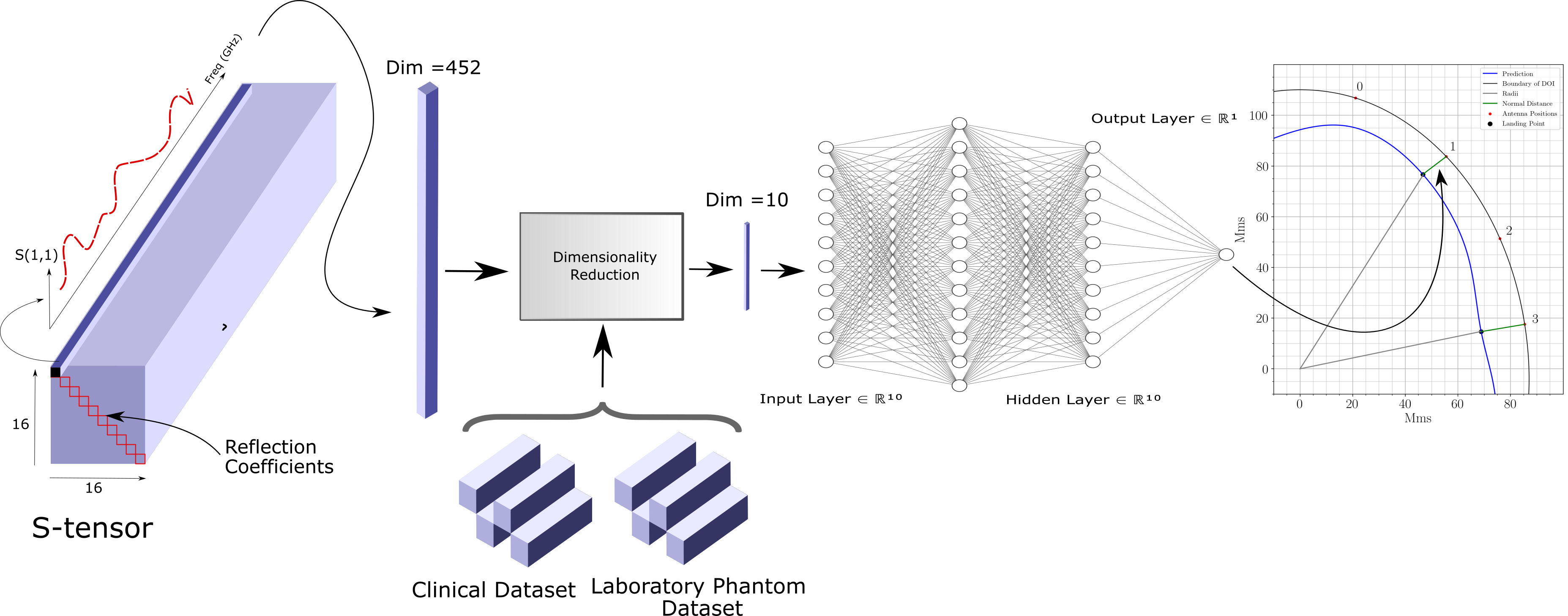}}
    \caption{\small The overall scheme of the proposed boundary detection algorithm}
    \label{fig:scheme}
\end{figure*}

\begin{itemize}
    \item Drastic reduction in potential over-fitting. This comes as a natural consequence of having a smaller model as the input is smaller in size (signal from one antenna as opposed to having all signals from the array as an input). A common plague of neural nets is memorization. With this architecture however, without indeed learning the underlying phenomenon, there is smaller chance that the neural net could ever perform well, even on training data.
    \item Ability to generalize to richer set of shapes beyond training data. Having built only one data-set using only one simple ellipse shaped phantom, yet, it can generalize to arbitrary shapes.
    \item Smaller sized dataset is sufficient to train the model. This is attributed to having a single model acting on consecutive antennas of a single measurement one by one. Thus, effectively multiplying the size of the dataset given by 16 (the number of antennas in the array used).
    \item Regressing normals to the antennas is a superior approach to the prediction of other response variables. This was shown to be the case indeed empirically, and is also physically explicable as it relates to the total distance travelled by the wave before hitting the first scatterer.
\end{itemize}

The final design scheme is depicted in Figure~\ref{fig:scheme}.

% \section{Experiments}
% \label{sec:experiments}
% We have experimented with a large number of architectures, but for the sake of brevity, we only include the results of the best model discussed in section earlier.

% We also experimented with data format fed to the neural net, and we report better performance when Cartesian format of complex data is used over polar one. The discontinuities in the phase information while un-warping complicates the learning.

% An ensemble of 30 neural nets is constructed (given the tiny size of individual components, the ensemble will not constitute a burden of any type)

\section{Results and Discussion}
\label{sec:results}

This section covers the end-to-end inner-working of the proposed model, then, the procurement of ground truths and evaluation metrics and concludes with the clinical results.

\subsection{From scattering parameters to prediction}
\begin{figure}[h!]
    \centering
    \includegraphics[width=0.4\textwidth]{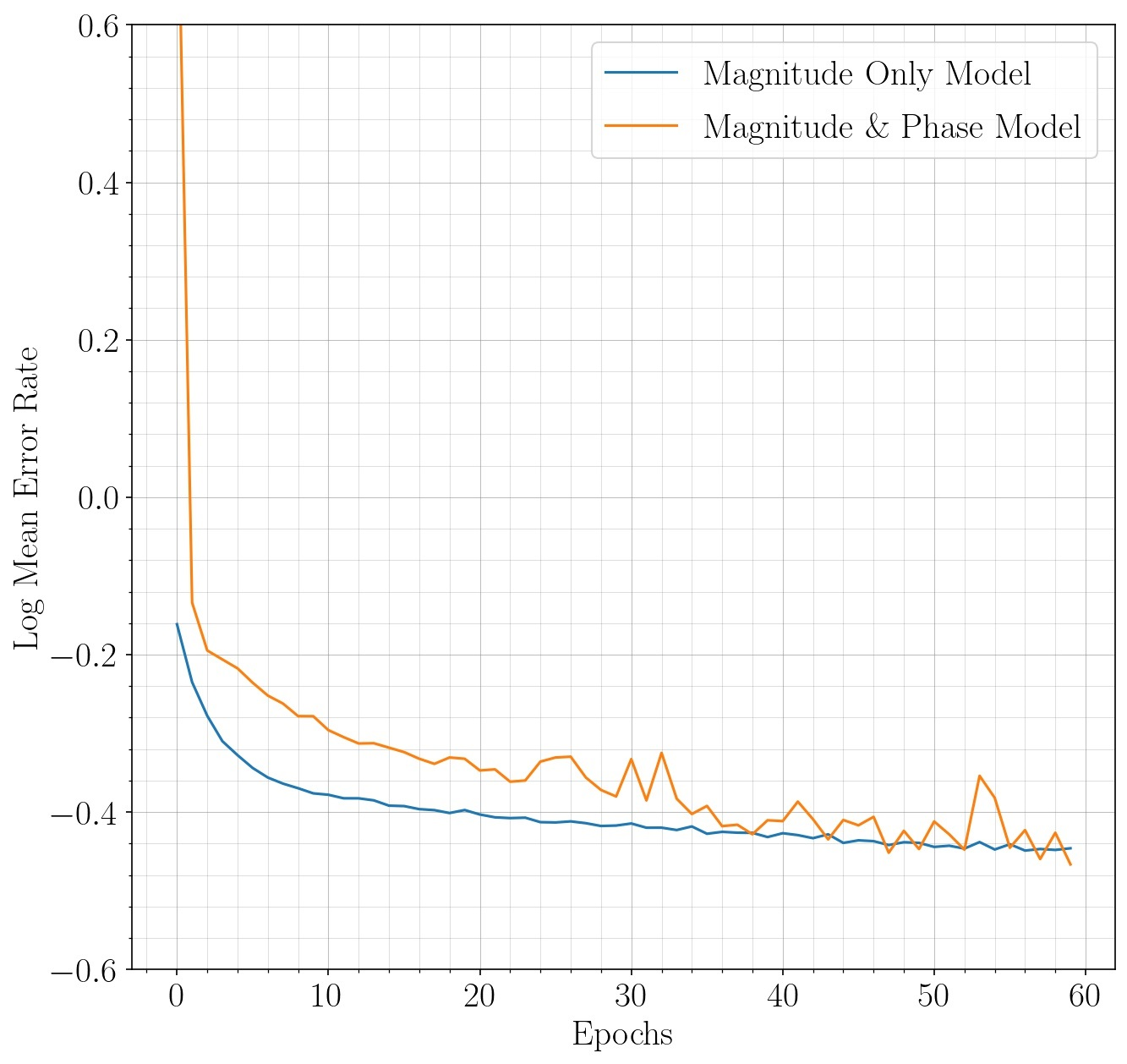}
    \caption{
    {\color{blue}
    Performance of magnitude-only model and complex-value on phantom training set.
    }
    }
    \label{fig:dtype}
\end{figure}

The training procedure for the  model depicted in Figure~\ref{fig:scheme} is delineated as follows:
% The model developed earlier in \ref{sec:ip_sig}, \ref{architect} \& \ref{resp_var} is trained in the following way:

\begin{itemize}
 \renewcommand{\labelitemi}{\scriptsize$\blacksquare$}
 \item  Preprocessing:\newline S parameters are not further calibrated beyond the calibration mentioned in Section \ref{sec:problem_setup}. 
 Given the band 0.7 -- 1.6~GHz, with the 2~MHz resolution, this results in 452 points in the signal.
 %Although increasing bandwidth helps, 
 %but in this case, it is known from the design stage that the antennas have reflection coefficients of greater than -10 dB outside the region above. (antennas misbehave)
 Then, PCA is used to fit the \textbf{magnitude} of the signal in the training set and eventually reduces it down to 10 dimensions. Notice that the fit results have to be saved and recalled at the deployment stage as it is part of algorithm inner-working. Additionally, scores of PCA are fed directly to the neural net and no further standardization is applied (otherwise it would make the neural net overly sensitive to the low ranking principal components).
 For the purpose of fitting PCA, 62 measurements (992 signals) from clinical settings were combined with only 2000 signals from training data which makes a reasonable balance between the two.
%  \newpage
%  {\color{blue}
 \item Model:\newline
 
 {\color{blue}
 The neural net itself is 4 layers deep, 10 neurons in width. LeakyRelu activations (with $\alpha=0.2$) are utilized after each layer apart from the last one which is activation-free. The model is compact and has a total of only 379 parameters. Figure~\ref{fig:hpttuning} depicts the hyperparameter search for the model with the least loss.
 }
 \begin{figure}[h!]
    \centering
    \includegraphics[width=0.4\textwidth]{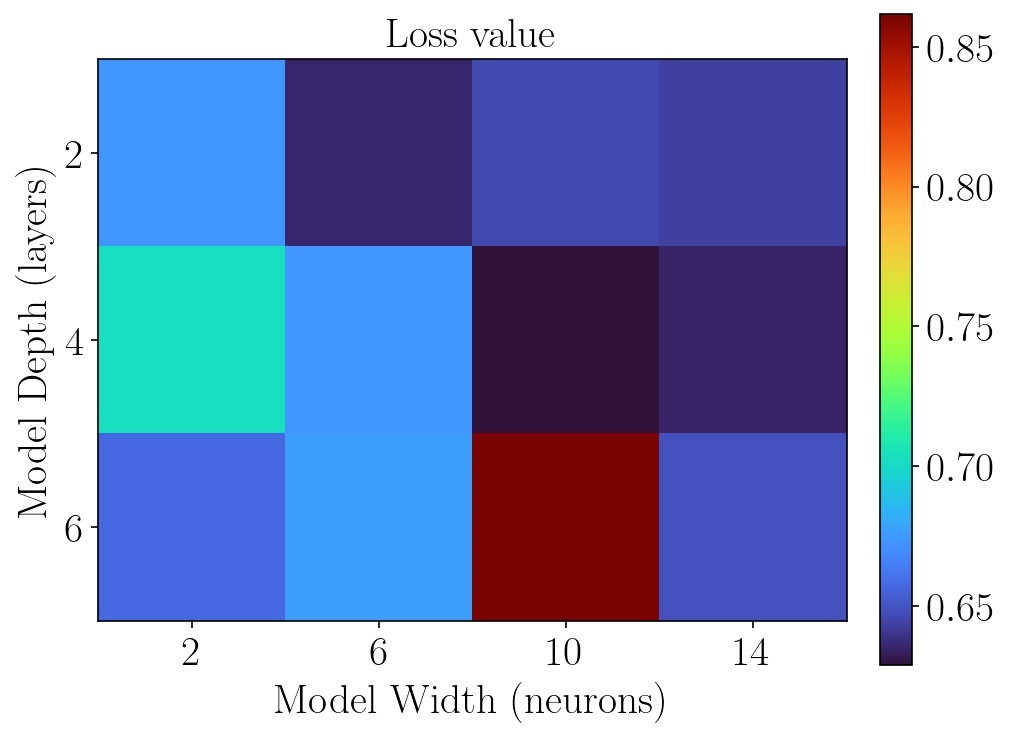}
    \caption{
    {\color{blue}
    Hyperparamer tuning for model width and depth.
    }
    }
    \label{fig:hpttuning}
\end{figure}

{\color{blue}

 \item Training:\newline
 Training with Adam optimizer and mean square error loss function was performed for 100 epochs with 32 batch size and a learning rate of 0.001. The splitting is performed such that 20\% of the data is reserved for testing. There were 444 \textit{measurements} (only training data come with labels that are useful for supervised training). However, given that a single measurement comprises of 16 antennas and predictions will be made per-antenna, dataset size becomes 7104.
% }
%  {\color{blue}
 The eminent compactness of the model incapacitates it to overfitting irrespective of how many epochs it is trained for. This thwarts the need for early stopping and regularizers that are usually used for large models that are heavy in the output and input data sizes.

 \item Postprocessing:\newline
 The estimated normals from the NN are scaled back and shifted to be on the millimeters scale as the training was performed on standardized labels. Following that, the landing points are calculated from the distances and antenna locations. Finally, cubic spline is applied to produce the final curve as discussed in \ref{resp_var}.
 
 }
\end{itemize}
{\color{blue}
Empirical results informed many of the design choices described above. Namely, the choice of the complex-value data type resulted in a model with almost twice the size (789 parameters), and marginally improved performance. Two PCA models were fit (one for each part of complex reflection coefficient), then, these values are concatenated and fed to a \textit{similar} model described earlier, apart from the first layer that is doubled in width to accommodate for the increase in input data size. Figure~\ref{fig:dtype} depicts the loss curves in each case. More importantly, the enhanced performance in training data is not helpful as it results in models `addicted' to this domain of laboratory data (which has little value per se) whereas the ultimate goal is the clinical data.
}

% Datatype

% insertion
\begin{figure}[!t]
    \centering
    \includegraphics[width=0.4\textwidth]{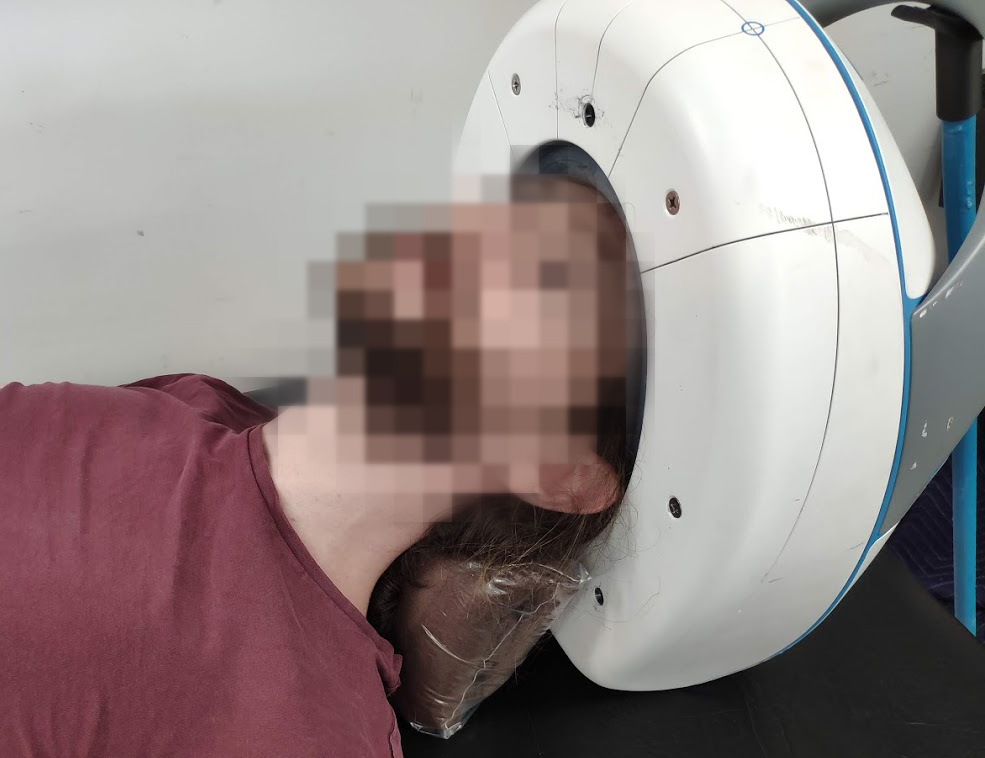}
    \caption{Modus operandi for patients EM scanning. A disposable surgical cap is worn and a support for the neck is provided while the head is inserted. Notice that the array is setup vertically in the clinical environment as opposed to the phantom-friendly horizontal setup shown in earlier in Figure~\ref{fig:experimental_setup}}
    \label{fig:insertion}
\end{figure}

% slices_test
\begin{figure*}[t!]
    \centering
    
    \begin{subfigure}[]{0.49\textwidth}
        \includegraphics[width=1.\textwidth]{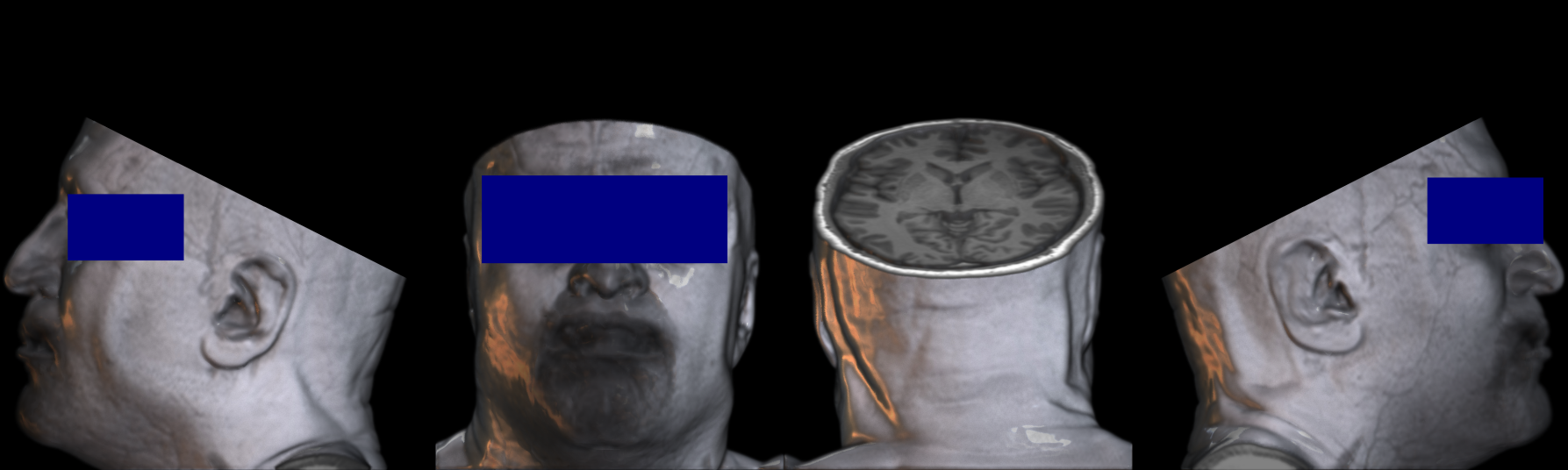}
        \caption{$depth=0.4, azimuth=0, elevation=70$}
        \label{fig:test_elev_70}
    \end{subfigure}
    \begin{subfigure}[]{0.46\textwidth}
        \includegraphics[width=1.\textwidth]{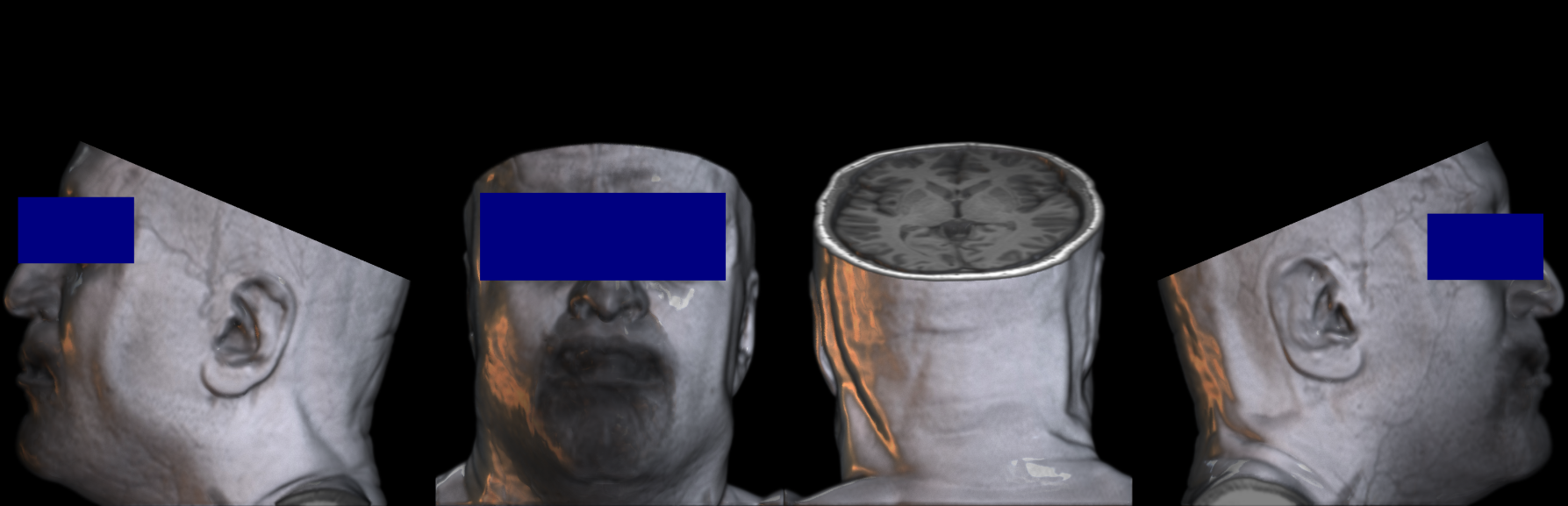}
        \caption{$depth=0.4, azimuth=0, elevation=65$}
        \label{fig:test_elev_65}
    \end{subfigure}

    \begin{subfigure}[]{0.28\textwidth}
        \includegraphics[width=1.\textwidth]{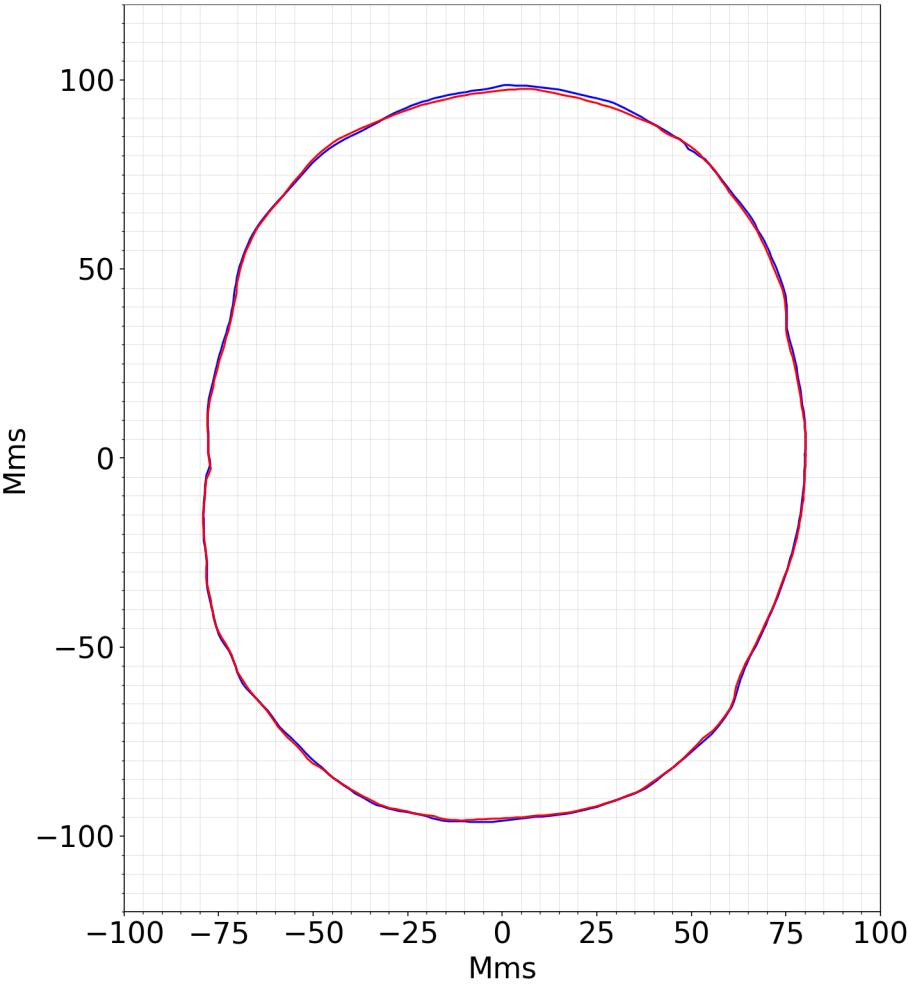}
        \caption{}
        \label{fig:elev_70}
    \end{subfigure}
    \begin{subfigure}[]{0.28\textwidth}
        \includegraphics[width=1.\textwidth]{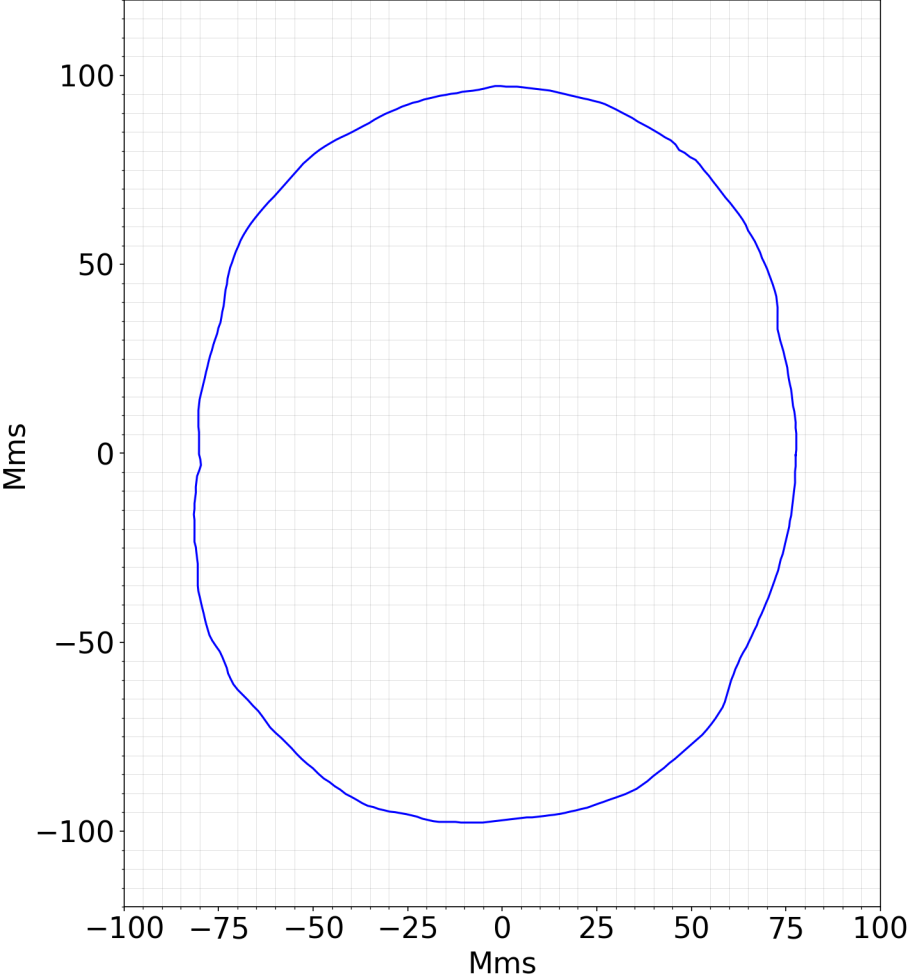}
        \caption{}
        \label{fig:elev_65}
    \end{subfigure}
    \begin{subfigure}[]{0.3\textwidth}
        \includegraphics[width=1.\textwidth]{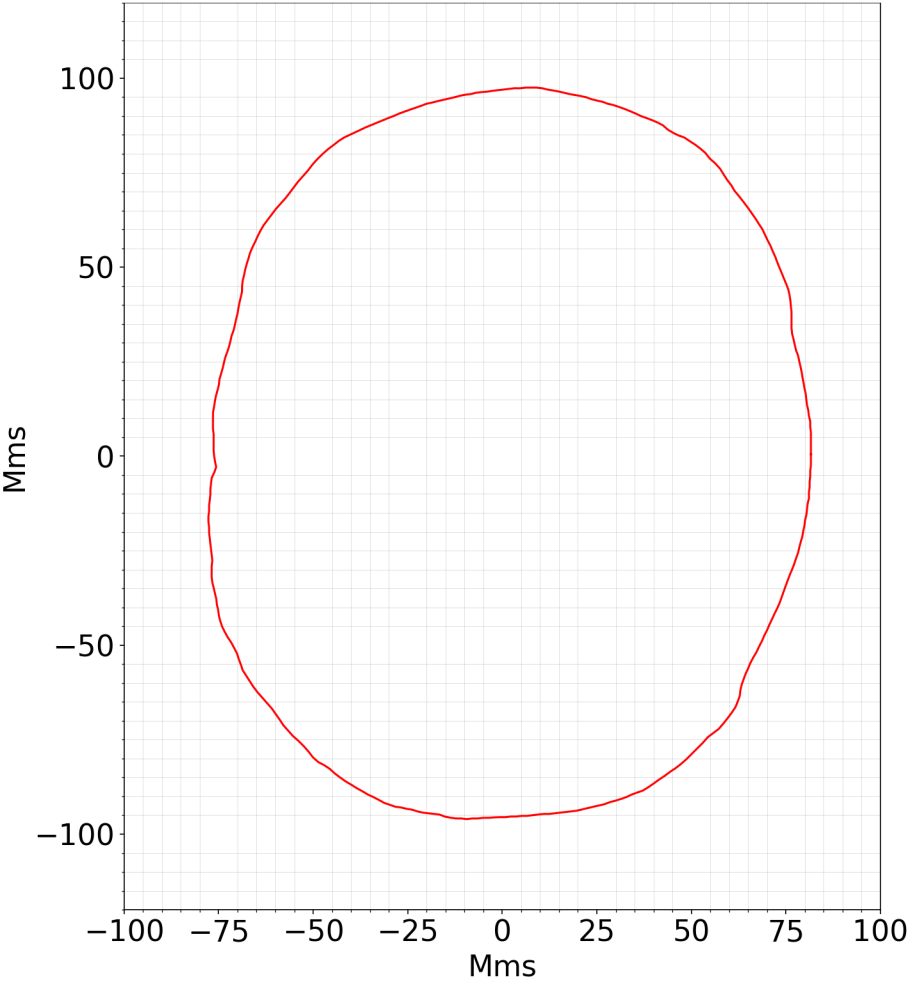}
        \caption{}
        \label{fig:test_elev_combined}
    \end{subfigure}
   
    \caption{Slice perturbation and boundary effect. (a) \& (b) show mosaics with different two different slices taken. (c) \& (d) show the corresponding boundaries respectively. Finally, (e) depicts the superimposition. Hu-difference is estimated for this case at $0.002$. Qualitatively, the maximum deviation between the two boundaries is around 1 mm.}  % should I say this explicitly?
    \label{fig:slices_test}
\end{figure*}

\subsection{Ground Truths \& Evaluation Metrics}
\label{sec:gt}
Procuring ground truths for algorithm evaluation when the subject is a human is arguably a technical difficulty. One might resort to photonic-based sensors, e.g. laser beams embedded in the array to estimate the normals. However, this approach is expensive and is severely impeded by non-transparent materials such as coupling medium or hair. 
On the other side, in electromagnetic based approach, the hair is invisible as it has low dielectric properties\footnote{Similarly in MRI, the hair is invisible as it has little if any water molecules. CT scans do not capture the hair vividly either.} \cite{martinsen1997dielectric}. 
A reasonable ground truth can be procured from conventional imaging modalities like MRI and CT. A simple boundary detection based on either CT or MRI of the same subject reveals the ground truth \textbf{shape} of the head, but not the \textbf{location} of the head inside imaging domain at EM scanning time. In one sense, it is not a golden ground truth, but it is sufficiently good for evaluation. The rationale behind this is that the only possible scenario in which the estimations by algorithm are wrong, yet, they match the shape of the ground truth, is when a constant bias error is made by the algorithm at individual antennas, and the negative of that constant bias is made at exactly at the corresponding diametrically opposite antenna. Evidently, such scenario has a negligibly minuscule chance of occurrence, thus, the evaluation is valid. As this holds true, it necessitates that when prediction and the ground truth are superimposed, only rigid shape transformations are allowed for the purpose of matching location.

The last vital piece of detail in ground truth procurement is the way in which the 3D MRI / CT scan is sliced to extract ground truth boundary. Needless to say, this slice should match the slice of head that was electromagnetically scanned. Antennas in the deployed array have a thickness of 3.4 centimeters, meaning that a \textit{thick} slice of the head is being scanned, rather than a 1 mm slice usually obtained from MRI/CT. Based on scanning protocol that was laid down, we specify that eyes and eyebrows are left outside and ears are left outside the setup as well, resulting in a slanted slice. This is both necessary, and useful as it allows scanning a wider slanted slice rather than transverse one, which is the case indeed in an air-based system that hangs from above and the patient is inserted from below. This scanning position described here is illustrated in Figure~\ref{fig:insertion}. This setup determines the slice being scanned to a great extent, and as such it serves as the basis for extracting the ground truth from an MRI scan. Additionally, several adjacent slices are taken to investigate how much of impact does this manual estimation affects the resulting boundary. The results are depicted in Figure~\ref{fig:slices_test}. It is noticeable that for 5 degrees difference in elevation, the boundaries match for the majority of points, while the maximum divergence is around \textbf{1 mm}. There is less change for perturbing the depth. This is not universally true; for other regions of the head this much perturbation results in a bigger change.

% res_vs_nn
\begin{figure*}[!t] %!tbp h
  \centering
  \subfloat[]{\includegraphics[width=0.4\textwidth]{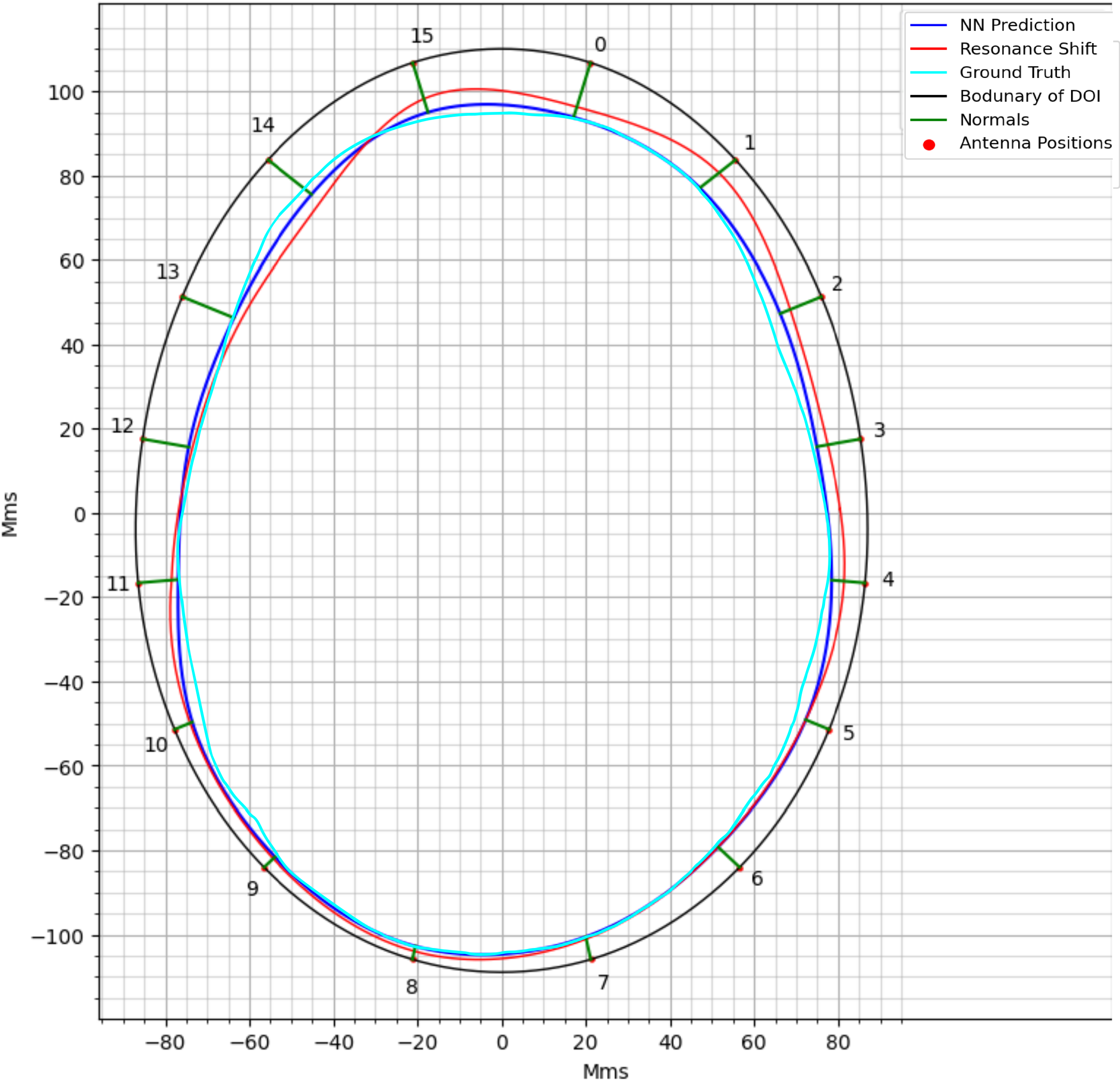}\label{fig:f1}}
%   \hfill
  \subfloat[]{\includegraphics[width=0.4\textwidth]{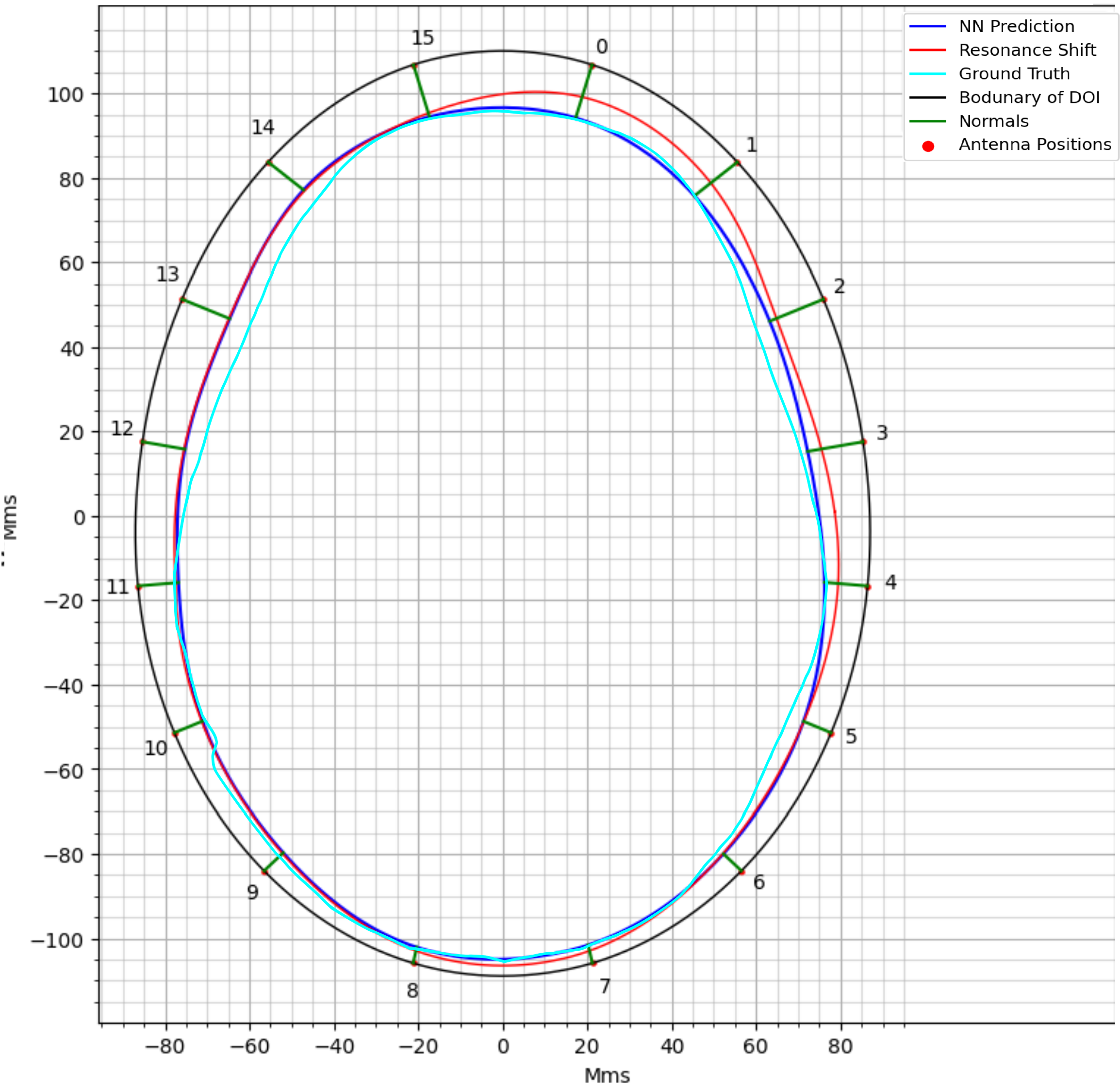}\label{fig:f2}}
  \caption{The proposed learning-based approach compared to the resonance-shift approach for two different cases (a) and (b). The portions of boundary where there is a match are due to coincidentally correct \textit{default} solutions picked by Resonance Shift.}  % those are cases 6 and 8. The title was removed because it is not meaningful for the the reader.
  % Both cases (a) \& (b) belong to human subjects from the clinical environment captured independently.
\label{fig:res_vs_nn}
\end{figure*}

%results
\begin{figure*}[!t]
    \centering
    
    \begin{subfigure}[]{0.48\textwidth}
        \includegraphics[width=1.\textwidth]{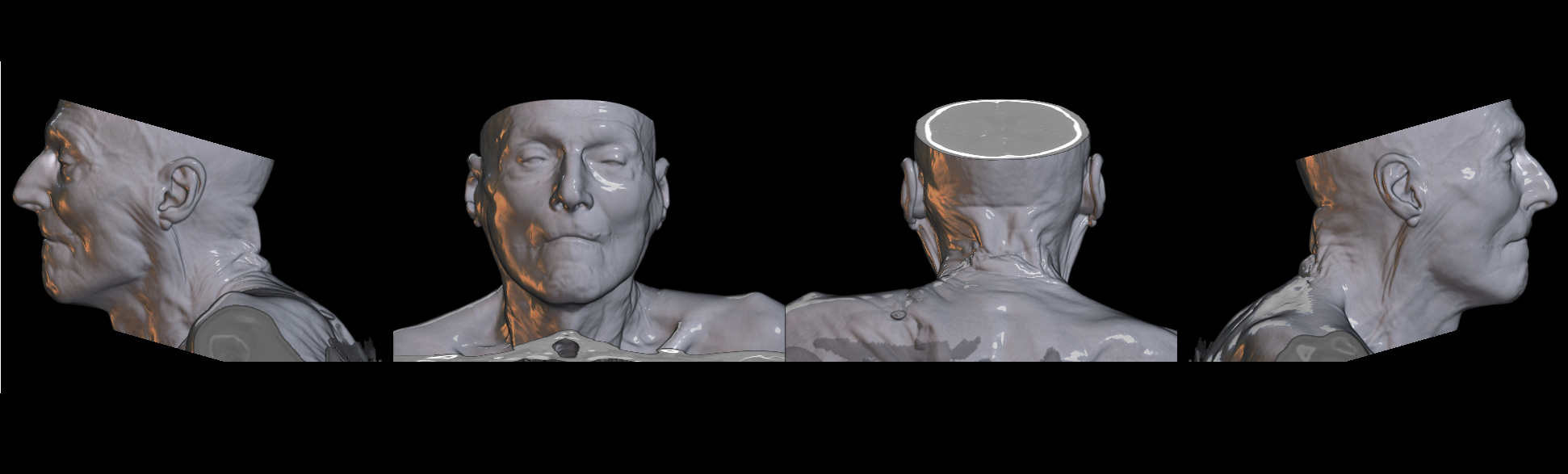}
        \caption{}
        \label{fig:p30_mosaic}
    \end{subfigure}
    \begin{subfigure}[]{0.15\textwidth}
        \includegraphics[width=1.\textwidth]{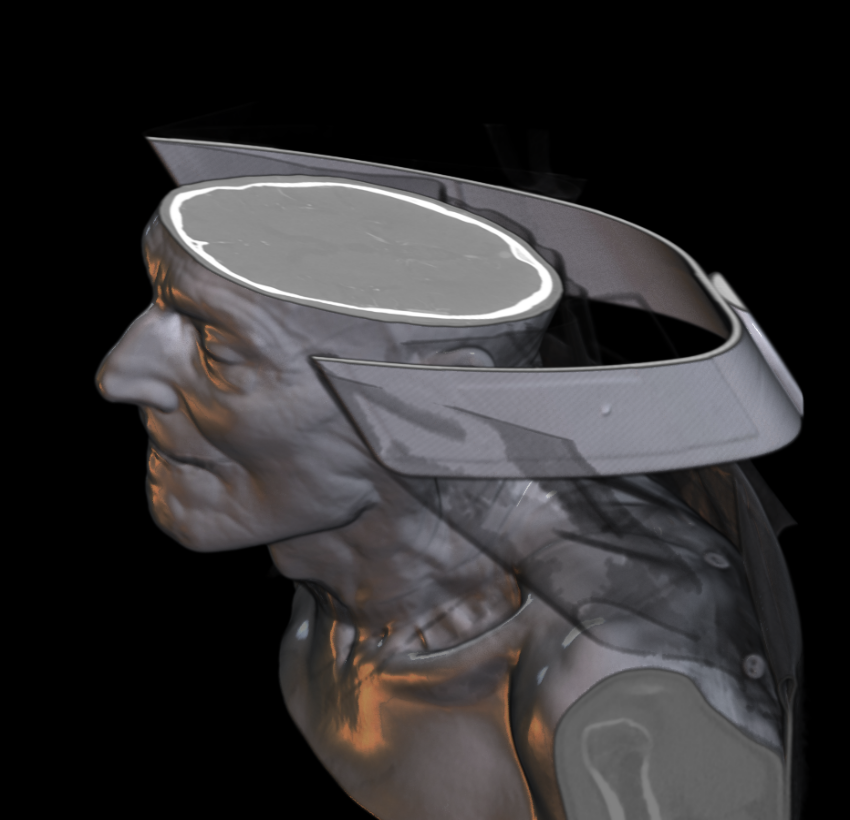}
        \caption{}
        \label{fig:default}
    \end{subfigure}    
    \begin{subfigure}[]{0.15\textwidth}
        \includegraphics[width=1.\textwidth]{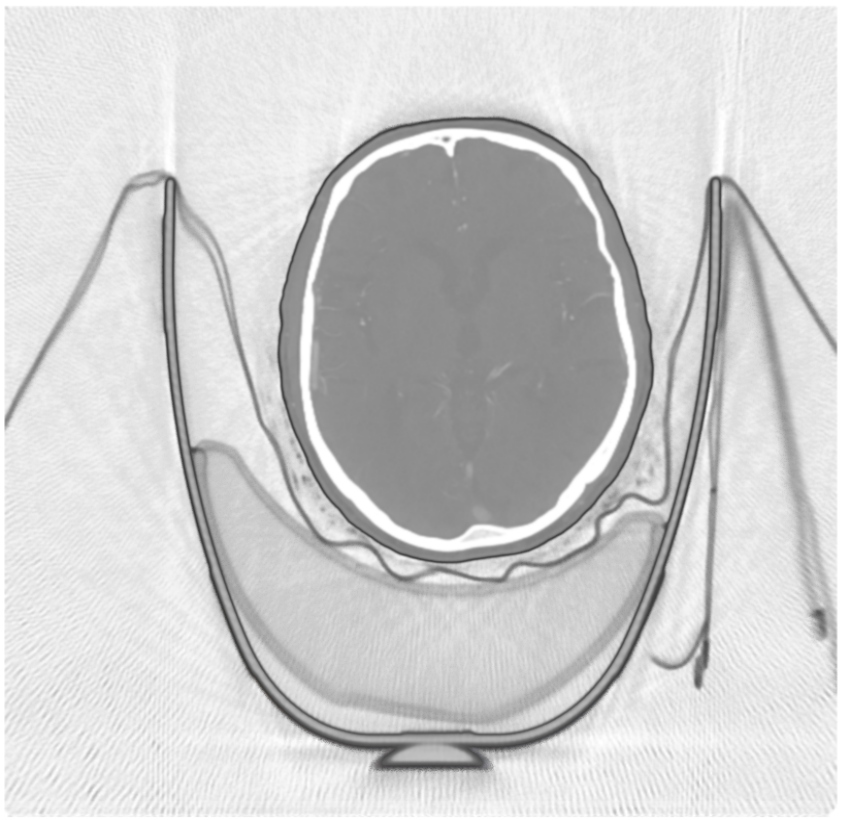}
        \caption{}
        \label{fig:hazed}
    \end{subfigure}  
    \begin{subfigure}[]{0.15\textwidth}
        \includegraphics[width=1.\textwidth]{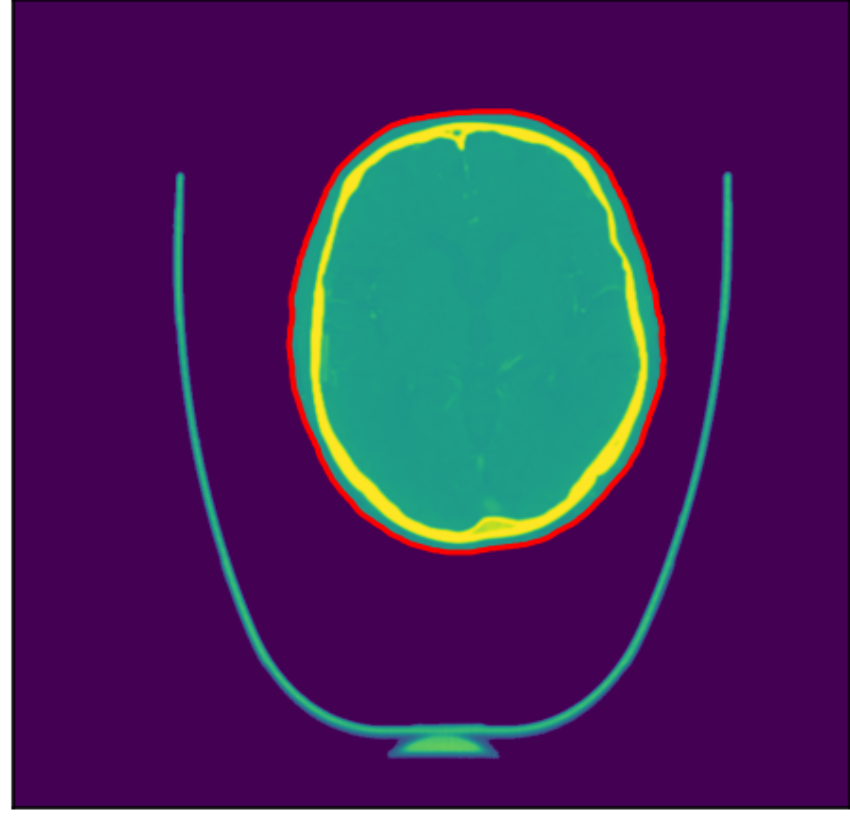}
        \caption{}
        \label{fig:boundary}
    \end{subfigure}  

    \begin{subfigure}[b]{0.3\textwidth}
        \includegraphics[width=1.\textwidth]{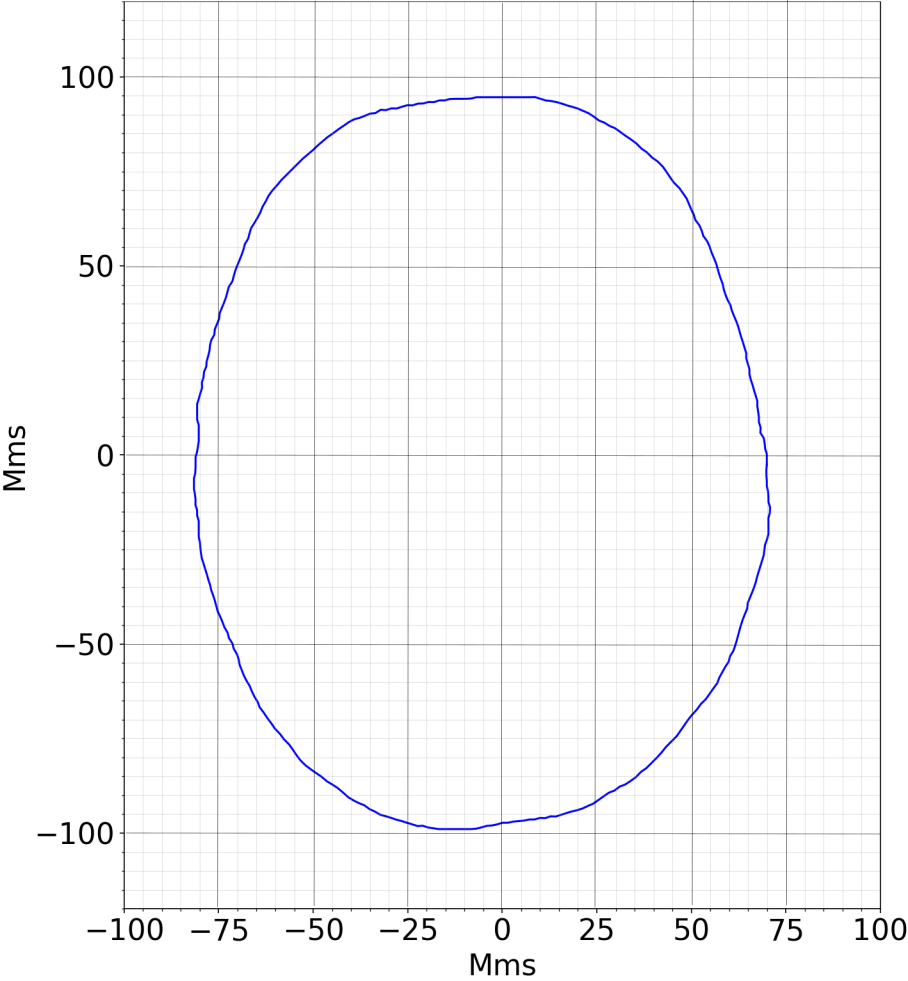}
        \caption{}
        \label{fig:1}
    \end{subfigure}  
    \begin{subfigure}[b]{0.3\textwidth}
        \includegraphics[width=1.\textwidth]{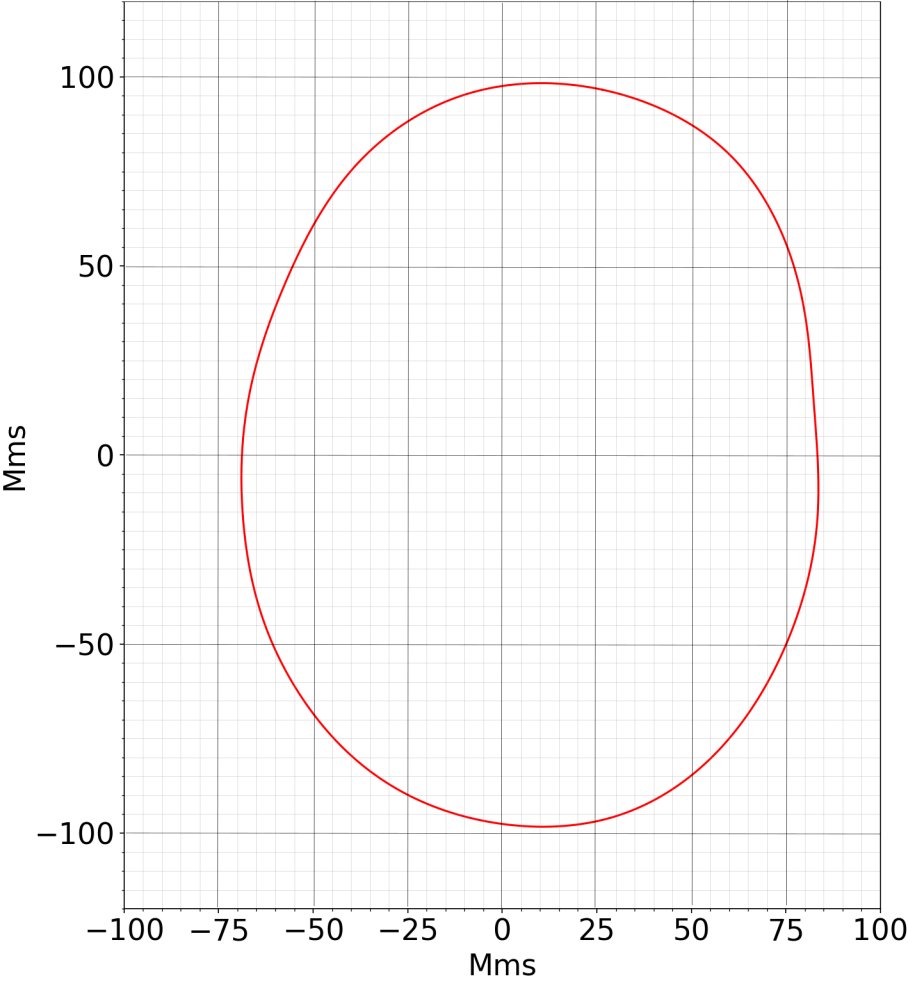}
        \caption{}
        \label{fig:2}
    \end{subfigure}  
    \begin{subfigure}[b]{0.3\textwidth}
        \includegraphics[width=1.\textwidth]{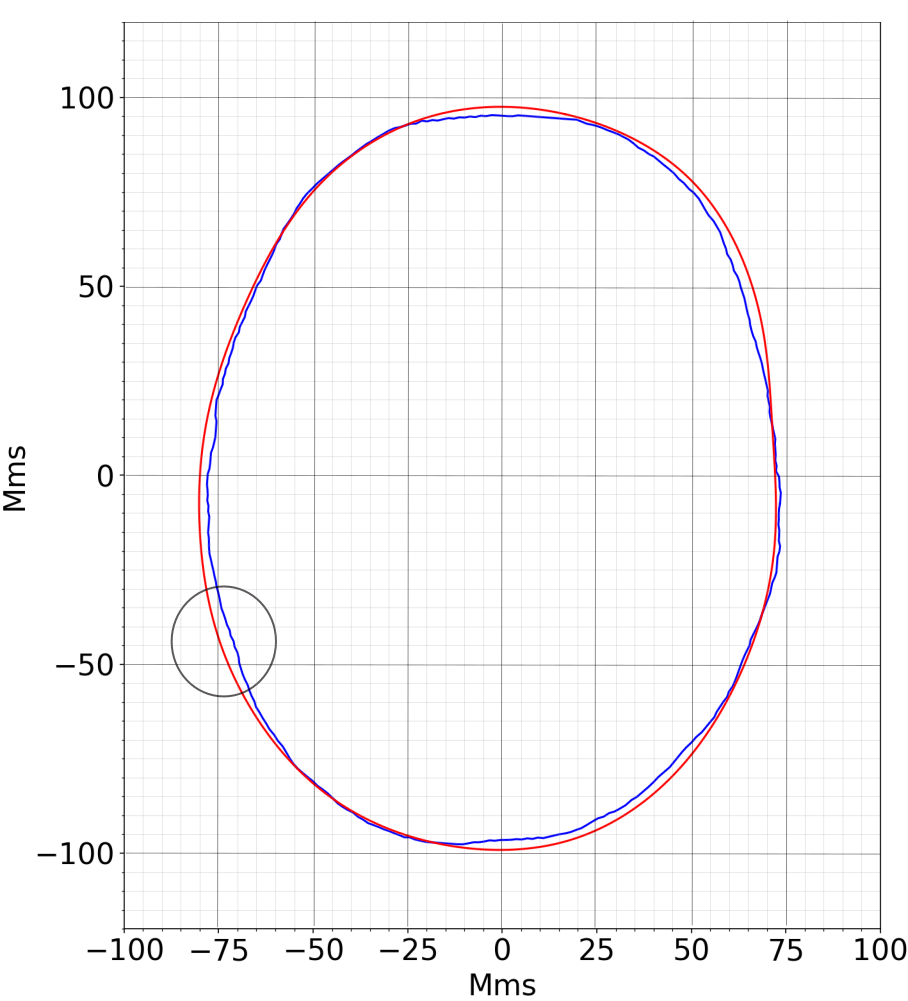}
        \caption{}
        \label{fig:3}
    \end{subfigure}  
        
\caption{(a) Mosaic of views that highlight the slice of interest. (b) A default view of the cut. (c). A perspective projection of the slice. (d) Image space boundary detected in the slice (shown in red). (e) Boundary extracted from CT and plotted in true millimeter scale grid. (f) Boundary as estimated by algorithm for the same subject. (g) Estimation and ground truth are overlapped. The circle in (g) shows a maximum difference of 3.5 mm.}  % should I say this explicitly?
\label{fig:results}
\end{figure*}

% example
\begin{figure*}[b!]
    \begin{subfigure}[b]{0.32\textwidth}
        \includegraphics[width=1.\textwidth]{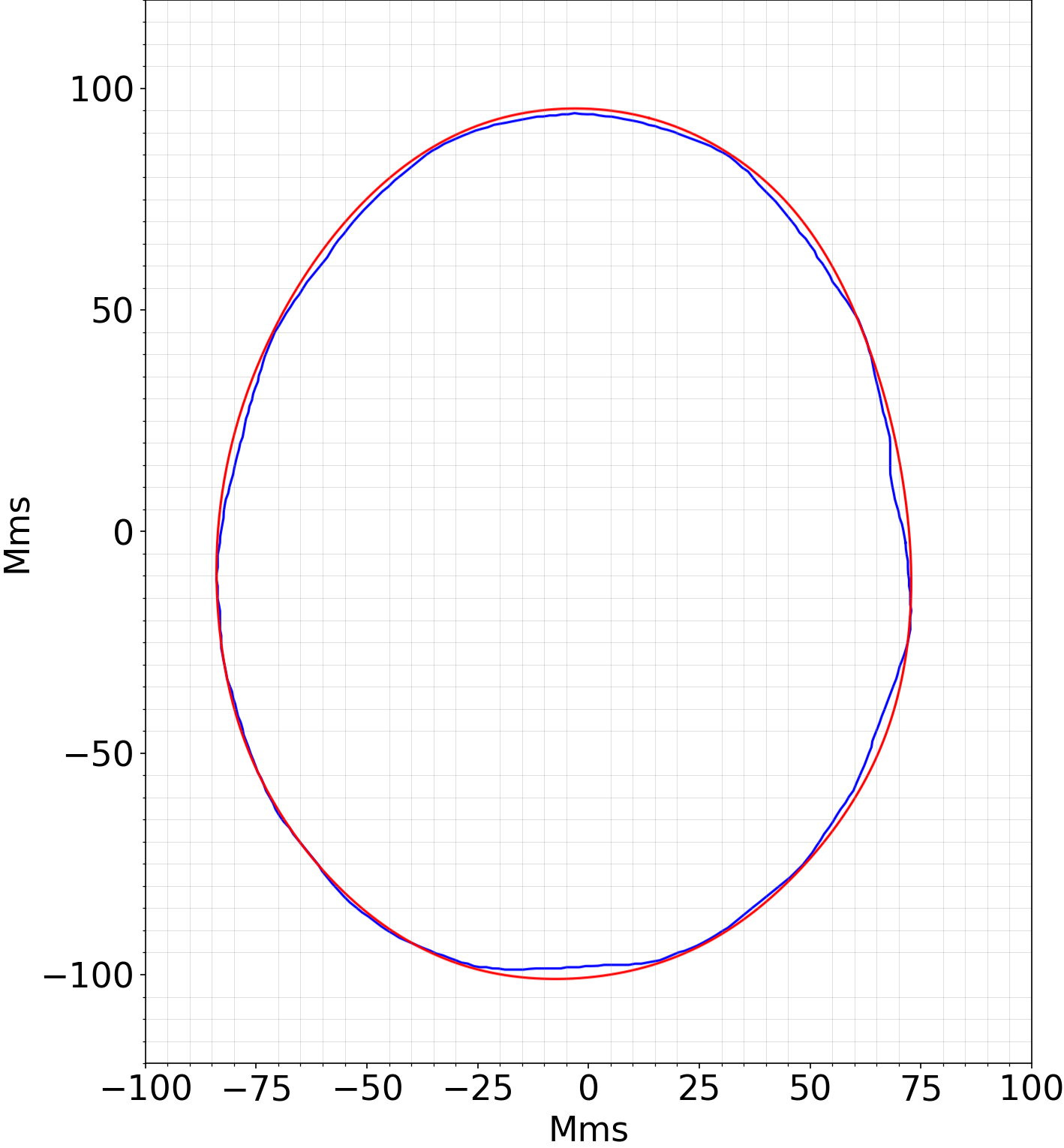}
        \caption{}
        \label{fig:ex1}
    \end{subfigure}  
    \begin{subfigure}[b]{0.32\textwidth}
        \includegraphics[width=1.\textwidth]{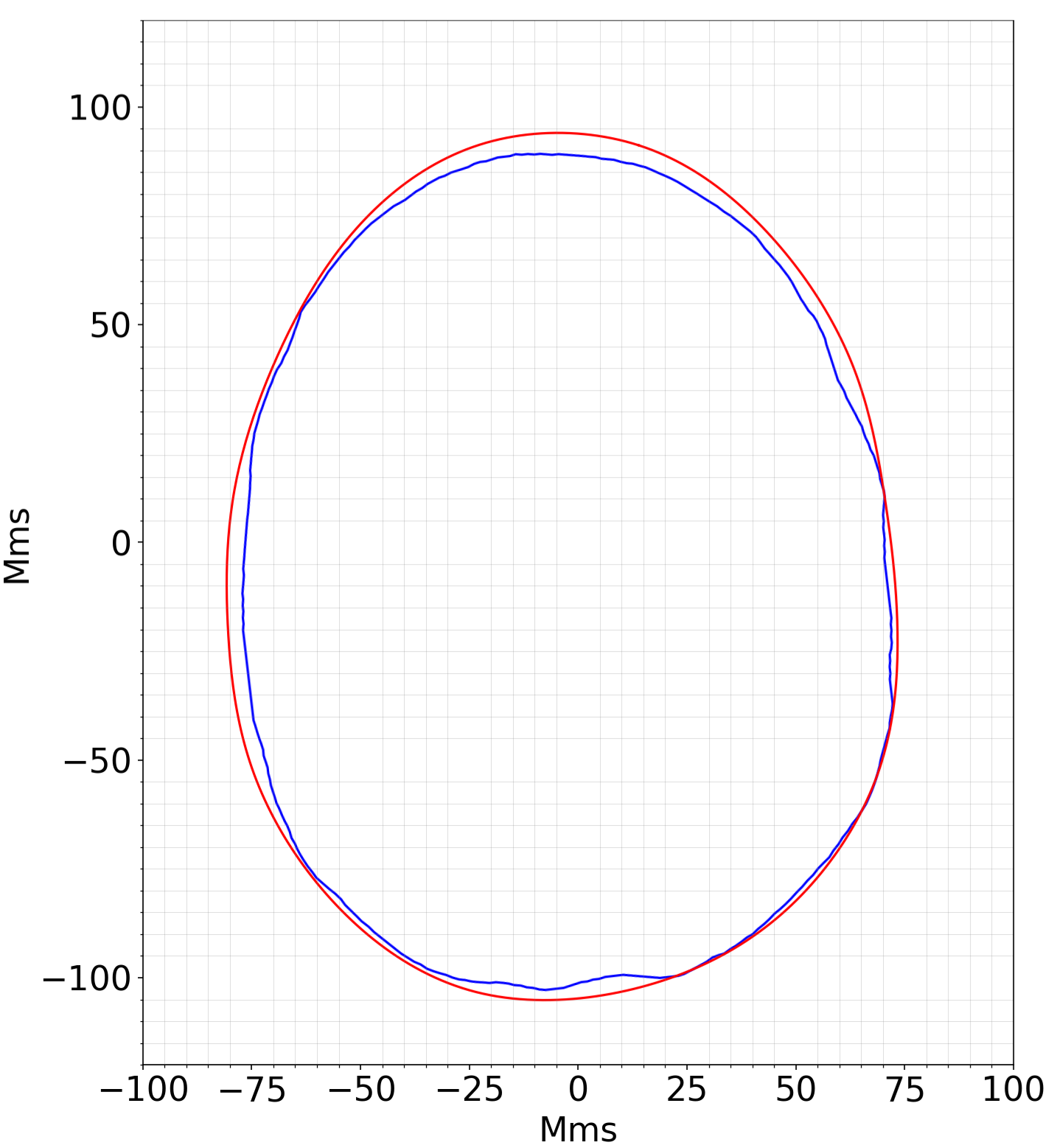}
        \caption{}
        \label{fig:ex2}
    \end{subfigure}  
    \begin{subfigure}[b]{0.32\textwidth}
        \includegraphics[width=1.\textwidth]{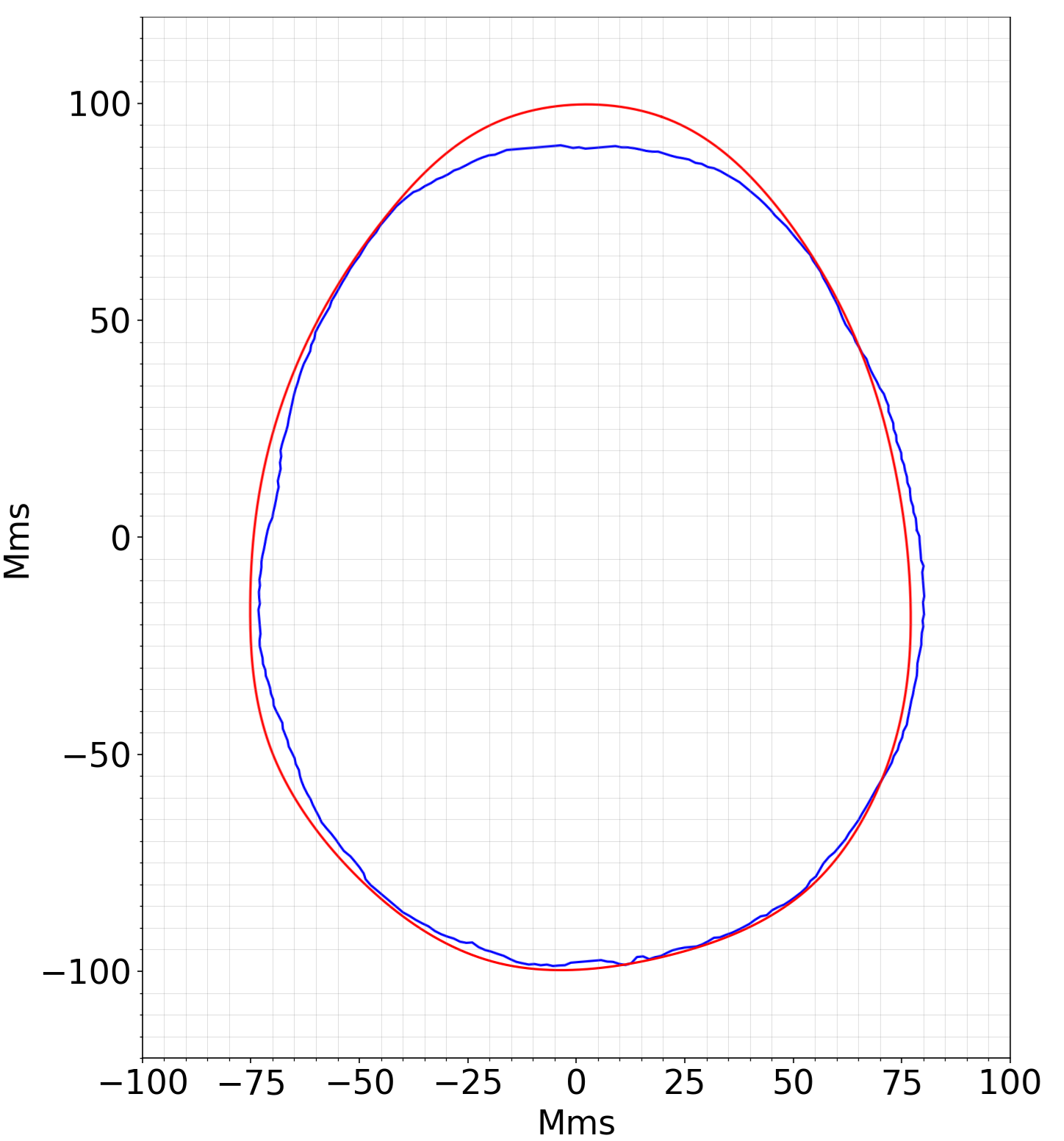}
        \caption{}
        \label{fig:ex3}
    \end{subfigure}  
    
\caption{Overlap of three ground truths (blue) with estimated boundaries from EM scans made in clinical environment (red). (a) showcases one of the best results. (b) shows a mediocre performance, while (c) depicts the result for a bad case.} 
\label{fig:three_cases}

    %The full slices of MRIs and intermediate results for each one of the cases is omitted for the sake of brevity. but provided in the supplementary material.}  % should I say this explicitly?
\end{figure*}

Both qualitative and quantitative measures are used to assess the predictions made. Qualitatively, the inferred and ground truth boundaries  are overlaid, then the maximum deviation between them is estimated. Quantitatively, several shape dissimilarity measures are utilized.
{\color{blue}
While a large number of measures exist in the literature  \cite{similaritymeasures1}, \cite{similaritymeasures2}, the measure that lends itself naturally is 
% {\color{blue}
Hu-Moments \cite{hu1962visual} as it presents seven expressions constructed in a way that is invariant to transformations of images. In each expression, central moments are used as building blocks. Later, the seven expressions are combined in the following way to produce the final measure \cite{hu_explanation}:
}
If $A$ and $B$ are image space objects to be compared, and $h_{i}^{A}$, $h_{i}^{B}$ are the corresponding Hu-moments, then:
$$
m_{i}^{A} = sign(h_{i}^{A}) \cdot \log h_i^{A}
$$
$$
m_{i}^{B} = sign(h_{i}^{B}) \cdot \log h_i^{B}
$$
$$
I(A, B) = \sum_{i=1}^{7}{\left|\frac{1}{m_{i}^{A}} - \frac{1}{m_{i}^{B}}\right|}
$$

In addition, simpler measures like curve length difference and area difference are used\footnote{These are not sound measures to quantify the difference between two sets, and as such, are only used ancillarily.}. Trivially, these measures are invariant to rotation and translation. In any of the aforementioned dissimilarity measures; the lower the better. A lower bound for the Hu-moments-based metric is zero, while the an upper bound does not exist as the two shapes can be arbitrarily dissimilar. 

\begin{table*}[]
\begin{tabular}{c|c|c|c|c|c|c|l}
\hline
Case & Hu (NN) & Hu (ReSh) & \% Area Change (NN) & \% Area Change (ReSh) & \% Length Change (NN) & \% Length Change (ReSh) & Gender \\ \hline
0    & 0.55    & 4.020     & 1.25                & 5.350                 & 0.29                  & 2.640                   & female \\ \hline
1    & 1.58    & 1.760     & 1.84                & 0.550                 & 1.57                  & 1.180                   & male   \\ \hline
2    & 0.71    & 0.290     & 2.12                & 5.900                 & 1.48                  & 2.420                   & male   \\ \hline
3    & 0.8     & 1.360     & 0.95                & 6.230                 & 0.88                  & 2.430                   & male   \\ \hline
4    & 0.71    & 2.800     & 7.03                & 11.310                & 4.28                  & 5.960                   & male   \\ \hline
5    & 1.53    & 0.160     & 1.59                & 1.150                 & 0.15                  & 0.060                   & male   \\ \hline
6    & 0.45    & 0.130     & 0.27                & 3.830                 & 0.92                  & 1.160                   & male   \\ \hline
7    & 2.57    & 3.810     & 15.63               & 2.200                 & 6.88                  & 0.020                   & male   \\ \hline
8    & 3.25    & 0.470     & 5.93                & 15.220                & 6.33                  & 8.540                   & male   \\ \hline
9    & 0.05    & 0.070     & 2.17                & 2.350                 & 0.61                  & 1.710                   & female \\ \hline
10   & 0.16    & 2.500     & 3.4                 & 4.230                 & 1.26                  & 2.110                   & female \\ \hline
11   & 2.4     & 1.890     & 5.19                & 17.500                & 2.3                   & 7.210                   & male   \\ \hline
12   & 0.72    & 2.500     & 6.0                 & 4.740                 & 2.57                  & 2.340                   & female \\ \hline
13   & 2.0     & 2.660     & 2.48                & 9.180                 & 0.21                  & 4.750                   & female \\ \hline
14   & 0.37    & 5.530     & 5.43                & 0.980                 & 2.03                  & 2.020                   & female \\ \hline
15   & 0.73    & 0.730     & 5.32                & 7.370                 & 3.32                  & 4.110                   & male   \\ \hline
16   & 0.2     & 1.900     & 2.87                & 3.700                 & 0.63                  & 0.950                   & male   \\ \hline
17   & 1.71    & 1.010     & 1.21                & 1.940                 & 1.66                  & 1.730                   & ?      \\ \hline
18   & 0.51    & 1.580     & 9.61                & 13.030                & 5.7                   & 6.810                   & male   \\ \hline
19   & 2.18    & 5.030     & 13.26               & 4.050                 & 8.18                  & 2.160                   & male   \\ \hline
\rowcolor[HTML]{EFEFEF} 
min  & 0.05    & 0.07      & 0.27\%              & 0.55\%                & 0.15\%                & 0.02\%                  &        \\ \hline
\rowcolor[HTML]{EFEFEF} 
max  & 3.25    & 5.53      & 15.63\%             & 17.5\%                & 8.18\%                & 8.54\%                  &        \\ \hline
\rowcolor[HTML]{EFEFEF} 
mean & 1.204   & 2.01      & 4.975\%             & 6.0405\%              & 2.708\%               & 3.01\%                  &        \\ \hline
\end{tabular}

\caption{
{\color{blue}
Summary of algorithm performance on 20 clinical cases with a comparison to resonance shift. The proposed technique is indicated with (NN) while resonance shift is indicated with (ReSh). Hu-difference reported in the table is scaled by 100 for convenience.
}
}
\label{table}

\end{table*}

\subsection{Clinical Results}
A reasonable estimation of the boundary is sought after. Concretely, an accuracy in the order of $\pm 1.5 mm$ is aimed for. To justify this, a dent in the head of over 1 mm can be created by simply pressing the head with fingers, this is particularly true around the temple where there exists a thicker layer of fat and skin. Alternatively, the head demorphs itself by that much due its own weight when lying. Adding to that the effect of the hair and other factors that only contribute to increasing the murky boundaries of the head. In fact, even when given MRI or CT, one still has to define a threshold and binarize the image to delineate the boundaries. Thus, one can only discuss the boundaries of the head up to this accuracy, below which the head is amorphous.  % Amin's comment: this is also a limitation of elemctromagnetic imaging.

% These results relate to laboratory experiments conducted by the research team. 
The results shown here are related to the clinical environment\footnote{Ethics Approval is recorded here: HREC/2019/QMS/48520} cases for which the scanned patients have accessible MRIs and / or CTs for ground truth comparison. The measurements were made independently by a clinician. The clinical data come from a different array of similar structure. 
%On paper, the two arrays are similar. In reality, 
Any discrepancies are attributed to manufacturing imperfections. (3C data vs 3B array)

\begin{figure*}[t!]
    \centering
    \includegraphics[width=1.0\textwidth]{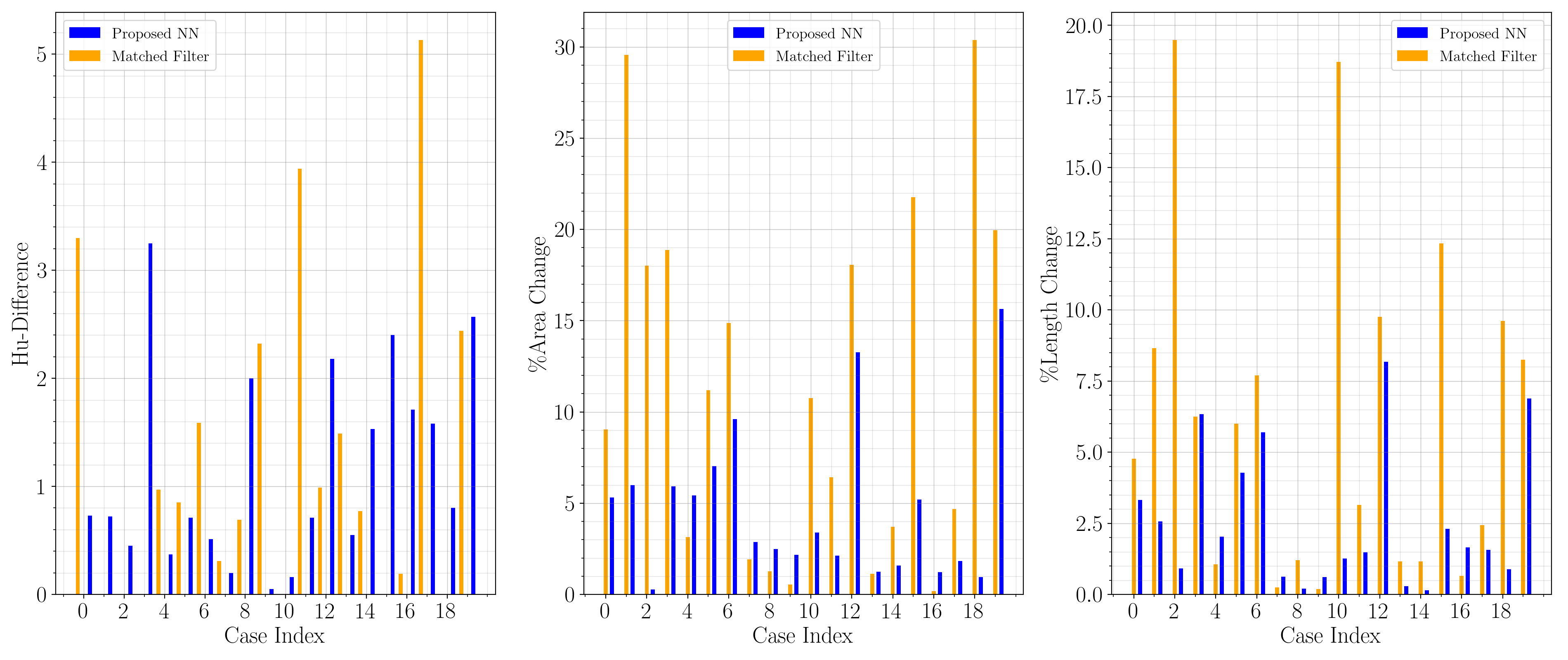}
    \caption{Comparison with Matched Filter technique as viewed by the three metrics indicated on Y-axis. Missing points are due to outlier estimations. In any of those metrics, lower difference from ground truth reflects better performance.}
    \label{fig:matched}
\end{figure*}

% \subsection{Resonance Shift VS Learning-based Model}
We start off by presenting outcome of the technique presented here as compared to the results generated by using the Resonance Shift method \cite{zamani2017boundary}. As discussed earlier, there is no unique result from Resonance Shift technique. In each antenna side, there exist possibly 3 different predictions. Thus, there will be a large combinatorial number of potential final shapes. Figure~\ref{fig:res_vs_nn} shows comparisons in two different cases. It ought to be emphasized that the ambiguity has not been resolved in those examples, but rather default predictions have been selected. The remaining results for this technique are omitted for this reason.
A note worth mentioning here, is that the resonance shift technique demands high frequency resolution at capture time to ensure accurate results. In particular, by inspecting the derivative of the plot in Figure~\ref{fig:resonance_trend} one can see a 2~MHz resolution to achieve 1 mm accuracy. This could be a tall order and increases acquisition time in an application where the object of interest is animate. ML approach on the other hand, doesn't seem to require this, at least not explicitly.
{\color{blue}
Concretely, the reflection coefficients signal of test set (which exhibit smooth behaviour) was sub-sampled to one eighth of its resolution then interpolated again, and fed to the model. The percentage increase in error rate was virtually non-existent 0.016\%. In comparison, resonance shift would quadruple its worst-case performance due to such reduction in resonance shift identification.
It must be emphasized here, that aside from this experiment, every other experiment conducted on all techniques are performed using the same input signals to ensure level-ground comparison.
}
% TODO: Should we complain about measurements not being perfect and there are hidden factors? if so they will demand experiments in which everything is under control.
% Ali: Not needed.
% TODO: Ali must tag his version of Resonance Shift algo to keep a record of results presented here.

% \subsection{Clinical Results}
The remaining results focus on evaluation of the proposed technique as compared to the ground truth.
% To present all-encompassing view of the results, three figures with different levels of details are presented. 
In Figure~\ref{fig:results}, the process of ground truth procurement is pictorially elaborated for a single clinical case. The sub-figures \ref{fig:1}, \ref{fig:2} and \ref{fig:3} show the ground truth, the inference and the superimposition respectively. While in Figure~\ref{fig:three_cases}, only the final superimposition for 3 more cases is presented. The cases are deliberately selected to reflect the overall performance. {\color{blue} Table~\ref{table} quantitatively describes the full clinical results. Inspection of the worst-case scenario (max), and average performance (mean) rows of the table, reveals a clear improvement of the proposed model over resonance shift method by all metrics.

A comparison is also conducted with Matched Filter technique from the literature \cite{Winters2008} which estimates delay in propagation till the first scatterer as described in the introduction. Again, a comfortable superiority is demonstrated by the learned method as depicted by Figure~\ref{fig:matched}. The implementation in \cite{Winters2008} implicitly requires a wide-band signal which would provide sufficiently high radar resolution. For the system at hand, only 1.5 GHz of bandwidth is utilized. The antennas comprising the array are purposefully designed to operate on frequencies that can penetrate the head for imaging.

}

% •	My attention was brought to a point he raised, which is the head having a pointy end of it resting on the antennas.
% o	This is a difference from the bucket that I never thought about.
% o	Yes SSB has good properties, but shape-wise, it is still a bucket.
% •	The hair is a point that missed raising. We don’t have a picture of the patient. Like very simple pictures could be immensely helpful in reasoning about predicted boundaries. I think Ali agrees with me.
% o	I will start looking at the gender which is given and see how if there’s any correlation there with the results.

The neural net seems to have learnt the underlying physical phenomenon. This was evidenced by its generalization ability when tested on clinical data. In particular, the least distance encountered in the training data was 3.83 mm, in practice, the neural net inferred distances below this (close to 1 mm). Similarly, generalization beyond the upper bound of training set was observed.

Accuracy-wise, not surprisingly, the performance was uneven. One can observe that estimations towards the bottom (where the pressure is higher) are very solid, while being more fragile at the top where the air bubbles inside the coupling medium usually settle. The hair doesn't seem to have a big impact as indicated by lack of correlation between performance metrics and gender in Table~\ref{table}. Needless to say, there are numerous hardware factors that play a role in determining the quality of the captured data, and subsequently the inference made on it, making it a challenge to isolate problems related to algorithm per se.

\section{Conclusions}
\label{sec:conclusions}
An operational data driven model from a feasible data collection mechanism was introduced. The  presented design enables successful learning of shape estimation from limited laboratory phantom data and generalizes to clinical data.
Performance-wise, the reported clinical results show that the goal of 1.5 mm accuracy was achieved in some cases but not all. The patchy performance is unsurprising given the high sensitivity of the data captured and the animate nature of the subject of interest. As such, a worthy future direction can be the design of an error correction mechanism on top of inference results. In practice, erroneous measurements are inevitable, resulting in erroneous predictions. Some notable reasons are air gaps between head and coupling medium, air bubbles inside coupling medium, a faulty antenna or a moving head. 
%The detection mechanism could be based on uncertainty estimates and the fix is based on .
The prior information, e.g. convexity or symmetries of shape to be estimated can serve as the basis for correction. 
%This has not been explicitly made use of in the scheme presented.
Such a mechanism increases the robustness of the algorithm, enabling it to produce reasonable results even in worst-case scenarios.

% \section{Supplement}
% Code to reproduce the results, architectures and fully trained models are available here:
% {\color{blue}\url{https://github.com/thisismygitrepo/boundary_estimation}}. 
% if have a single appendix:
%\appendix[Proof of the Zonklar Equations]
% or
%\appendix  % for no appendix heading
% do not use \section anymore after \appendix, only \section*
% is possibly needed

% use appendices with more than one appendix
% then use \section to start each appendix
% you must declare a \section before using any
% \subsection or using \label (\appendices by itself
% starts a section numbered zero.)
%

% \appendices
% \section{Proof of the First Zonklar Equation}
% Appendix one text goes here.

% % you can choose not to have a title for an appendix
% % if you want by leaving the argument blank
% \section{}
% Appendix two text goes here.

% use section* for acknowledgment
% \section*{Acknowledgment}

% The authors would like to thank...

% Can use something like this to put references on a page
% by themselves when using endfloat and the captionsoff option.
\ifCLASSOPTIONcaptionsoff
  \newpage
\fi

% trigger a \newpage just before the given reference
% number - used to balance the columns on the last page
% adjust value as needed - may need to be readjusted if
% the document is modified later
%\IEEEtriggeratref{8}
% The "triggered" command can be changed if desired:
%\IEEEtriggercmd{\enlargethispage{-5in}}

% references section

% can use a bibliography generated by BibTeX as a .bbl file
% BibTeX documentation can be easily obtained at:
% http://mirror.ctan.org/biblio/bibtex/contrib/doc/
% The IEEEtran BibTeX style support page is at:
% http://www.michaelshell.org/tex/ieeetran/bibtex/
%\bibliographystyle{IEEEtran}
% argument is your BibTeX string definitions and bibliography database(s)
%\bibliography{IEEEabrv,../bib/paper}
%
% <OR> manually copy in the resultant .bbl file
% set second argument of \begin to the number of references
% (used to reserve space for the reference number labels box)

\bibliographystyle{IEEEtran}
\bibliography{library.bib}

% biography section
% 
% If you have an EPS/PDF photo (graphicx package needed) extra braces are
% needed around the contents of the optional argument to biography to prevent
% the LaTeX parser from getting confused when it sees the complicated
% \includegraphics command within an optional argument. (You could create
% your own custom macro containing the \includegraphics command to make things
% simpler here.)
%\begin{IEEEbiography}[{\includegraphics[width=1in,height=1.25in,clip,keepaspectratio]{mshell}}]{Michael Shell}
% or if you just want to reserve a space for a photo:

% \begin{IEEEbiography}{Ahmed Al-Saffar}
% Biography text here.
% \end{IEEEbiography}

% % if you will not have a photo at all:
% \begin{IEEEbiographynophoto}{Ali Zamani}
% Biography text here.
% \end{IEEEbiographynophoto}

% % insert where needed to balance the two columns on the last page with
% % biographies
% %\newpage

% \begin{IEEEbiographynophoto}{Amin Abbosh}
% Biography text here.
% \end{IEEEbiographynophoto}

% You can push biographies down or up by placing
% a \vfill before or after them. The appropriate
% use of \vfill depends on what kind of text is
% on the last page and whether or not the columns
% are being equalized.

%\vfill

% Can be used to pull up biographies so that the bottom of the last one
% is flush with the other column.
%\enlargethispage{-5in}
\end{document}